\documentclass[review]{elsarticle}

\usepackage{lineno,hyperref}
\modulolinenumbers[5]

\journal{Applied and Computational Harmonic Analysis}
\usepackage{amsmath,amsfonts,amsthm,mathrsfs,xcolor, bm,amssymb}
\usepackage{verbatim}
\usepackage{graphicx}
\usepackage[export]{adjustbox}
\usepackage{color}
\usepackage{cancel}

\numberwithin{equation}{section}
\newtheorem{theorem}{Theorem}
\newtheorem{lemma}{Lemma}
\newtheorem{proposition}{Proposition}

\newtheorem{definition}{Definition}
\newtheorem{remark}{Remark}

\usepackage[scale={0.72,0.743}, hmarginratio=1:1, vmarginratio=1:1, headheight=2ex, headsep = 2ex, footskip = 3.9ex, nomarginpar]{geometry}

\def\NN{\mathbb N}
\def\ZZ{\mathbb Z}
\def\RR{\mathbb R}
\def\CC{\mathbb C}

\begin{document}
\begin{frontmatter}

\title{Learning Ability of Interpolating  Deep Convolutional Neural Networks}

\author{Tian-Yi Zhou}\corref{mycorrespondingauthor}
\cortext[mycorrespondingauthor]{Corresponding author}
\ead{tzhou306@gatech.edu}
\author{Xiaoming Huo}



\address{H. Milton Stewart School of Industrial and Systems Engineering, Georgia Institute of Technology}
\address{765 Ferst Drive,
Atlanta GA 30332-0205}

\begin{abstract}
It is frequently observed that overparameterized neural networks generalize well.
Regarding such phenomena,  existing theoretical work mainly devotes to linear settings or fully-connected neural networks.
This paper studies the learning ability of an important family of deep neural networks, deep convolutional neural networks (DCNNs), under both underparameterized and overparameterized settings.
We establish the first learning rates of underparameterized DCNNs without parameter or function variable structure restrictions presented in the literature.
We also show that by adding well-defined layers to a non-interpolating
DCNN, we can obtain some interpolating DCNNs that maintain the good learning rates of the non-interpolating DCNN.
This result is achieved by a novel network deepening scheme designed for DCNNs.
Our work provides theoretical verification of how overfitted DCNNs generalize well.
\end{abstract}


\end{frontmatter}


\section{Introduction}
Neural networks are computing systems with powerful applications in many disciplines, such as data analysis and pattern and sequence recognition.
In particular, deep neural networks with well-designed structures, numerous trainable free parameters, and massive-scale input data have outstanding performances in function approximation \cite{Yarosky, Klusowski2018}, classification \cite{Krizhevsky2012, caragea2020neural}, regression \cite{Schmidt-Hieber}, and feature extraction \cite{mao1995artificial}.
The success of deep neural networks in practice has motivated research activities intended to rigorously explain their capability and power, in addition to the literature on shallow neural networks \cite{Mhaskar1993}.

In this paper, we study an important family of neural networks known as convolutional neural networks. Given that neural networks, in general, are powerful and versatile, researchers have been working to improve their computational efficiency further. When the data dimension is large such as the AlexNet \cite{Krizhevsky2012} of input dimension about $150,000$, fully-connected neural networks are not feasible.
Structures are often imposed on neural networks to reduce the number of trainable free parameters and get feasible deep learning algorithms for various practical tasks \cite{Lecun1998}.
The structure we are interested in is induced by one-dimensional convolution (1-D convolution), and the resulting networks are deep convolutional neural networks (DCNNs) \cite{Waibel}.
The convolutional structure of DCNNs reduces the computational complexity and is believed to capture local shift-invariance properties of image and speech data.
Such features of DCNNs contribute to the massive popularity of DCNNs in image processing and speech recognition.

In recent years, there has been a line of work studying overparameterization in deep learning. It is frequently observed that overparameterized deep neural networks, such as DCNNs, generalize well while achieving zero training error \cite{belkin2018understand}. This phenomenon, known as benign overfitting, seems to confront the classical bias-variance trade-off in statistical theory. Such a mismatch between observations and classical theory sparked avid research attempting to understand how benign overfitting occurs. Theoretical work studying benign overfitting was initiated in \cite{BartlettBenign}, where a linear regression setting with Gaussian data and noise was considered. It presented conditions for minimum-norm interpolators to generalize well. In a non-linear setting induced by the ReLU activation function, benign overfitting is verified for deep fully-connected neural networks in \cite{lin2021generalization}. On top of that, a recent work \cite{cao2022benign} shows that training shallow neural networks with shared weights by gradient descent can achieve an arbitrarily small training error.

%

In this paper, we study the learning ability of DCNNs under both underparameterized and overparameterized settings. We aim to show that an overparameterized DCNN can be constructed to have the same convergence rate as a given underparameterized one while it perfectly fits the input data. In other words, we intend to prove that interpolating DCNNs generalize well.

The main contributions of the paper are as follows. Our first result rigorously proves that for an arbitrary DCNN with good learning rates,
we can add more layers to build overparameterized DCNNs satisfying the interpolation condition while retaining good learning rates.
Here, ``learning rates" refers to rates of convergence of the output function to the regression function in a regression setting. Our second result establishes the learning rates of DCNNs in general.
Previously in \cite{zhou2020universality}, convergence rates of approximating functions in some Sobolev spaces by DCNNs were given without generalization analysis.
Moreover, learning rates of DCNNs for learning radial functions were given in \cite{mao2021theory}, where the bias vectors and filters are assumed to be bounded, with bounds depending on the sample size and depths. More recently, learning rates for learning additive ridge functions were presented in \cite{fang2022optimal}.
Unlike these existing works, the learning rates we derive do not require any restrictions on norms of the filters or bias vectors, or variable structures of the target functions. 
Without the boundedness of free parameters, the standard covering number arguments do not apply.
To overcome such a challenge, we derive a special estimate of the pseudo-dimension of the hypothesis space generated by a DCNN. Previously, a pseudo-dimension estimate was given in  \cite{Bartlett2019} for fully-connected neural networks, using the piecewise polynomial property of the activation function. We shall apply our pseudo-dimension estimate to, in turn, bound the empirical covering number of the hypothesis space. In such a way, we can achieve our results without restrictions on free parameters.

Combining our first and second results, we prove that for any input data, there exist some overparameterized DCNNs which interpolate the data and achieve a good learning rate. 
The third result provides theoretical support for the possible existence of benign overfitting under the DCNN setting.

The rest of this paper is organized as follows.
In Section \ref{formulation}, we introduce notations and definitions used throughout the paper, including the definition of DCNNs to be studied.  
In Section \ref{results}, we present our main results that describe how a DCNN achieves benign overfitting. 
The proof of our first result is given in Section \ref{Proof1}, and the proofs of our second and third results are provided in Section \ref{Proof2}. 
In Section \ref{experiment}, we present the results of numerical experiments which corroborate our theoretical findings.
Lastly, in Section \ref{sec:discussion}, we present some discussions and compare our work with the existing literature.

\section{Problem Formulation} \label{formulation}
In this section, we define the DCNNs to be studied in this paper and the corresponding hypothesis space (Subsection \ref{sec:dcnn-def}). Then, we introduce the regression setting with data and the regression function (Subsection \ref{sec:data}). 
\subsection{Deep Convolutional Neural Networks and the Corresponding Hypothesis Space}
\label{sec:dcnn-def}
To begin with, we formulate the 1-D convolution.
Let $w=\{w_j\}_{j=-\infty}^{+\infty}$ be a filter supported in $\{0, 1, \ldots, s\}$ for some filter length $s\in \NN$, which means $w_j \neq 0$ only for $0\leq j\leq s$.
Suppose $x=\{x_j\}_{j=-\infty}^{+\infty}$ is another sequence supported in $\{1, 2, \ldots, d\}$ for some $d\in \NN$ and is denoted as an input vector $x=(x_1,\ldots, x_d)^T \in\RR^d$ for networks in the following.
The 1-D convolution of $w$ with $x$, denoted by $w{*}x$, is defined as
\begin{equation} \label{1Dconvolution}
 \left(w{*} x\right)_i = \sum_{k\in\ZZ} w_{i-k} x_k = \sum_{k=1}^d w_{i-k} x_k, \qquad i\in \ZZ.
\end{equation}
We can see, from (\ref{1Dconvolution}), the  convoluted sequence $w{*}x$ is supported in $\{1, 2, \ldots, d+s\}$ and can be expressed in the following matrix form
\begin{equation}
\left[\left(w{*}x\right)_i\right]_{i=1}^{d+s} =T^w x,
\end{equation}
where 
\begin{align} \label{convolution}
(T^w)_{i,i}&=w_0, \text{ if } i=1,2,\ldots, d, \nonumber\\
(T^w)_{i+1,i}&=w_1, \text{ if } i=1,2,\ldots, d,\nonumber\\
&\vdots\\
(T^w)_{i+s,i}&=w_s, \text{ if } i=1,2,\ldots, d, \nonumber\\
(T^w)_{i,j}&=0, \text{ otherwise.}\nonumber
\end{align}
$T^w$ is a $(d+s) \times d$ sparse Toeplitz-type matrix often referred as the ``convolutional matrix."
The sparsity of $T^w$ can be attributed to the large number of zero entries.
This approach is known as ``zero-padding," where we have expanded the vector $x\in\RR^d$ to a sequence on $\ZZ$ by adding zero entries outside the support $\{1, 2, \ldots, d\}$.

Now we define DCNNs by means of convolutional matrices.
We take the ReLU activation function $\sigma: \RR \to \RR$ given by $\sigma (u)= \max \{0,u\}$.
It acts on vectors componentwise.

\begin{definition}\label{defDCNN}
A DCNN of depth $J \in \NN$ consists of a sequence of function vectors $\{h^{(j)}: \RR^d \to \RR^{d_j}\}_{j=1}^J$ of widths $\{d_j :=d+j s\}_{j=0}^J$ defined
with a sequence of filters ${\bf w}=\{w^{(j)}\}_{j=1}^{J}$ each of filter length $s\in \NN$ and a sequence of  bias vectors ${\bf b} =\{b^{(j)} \in\RR^{d_j}\}$ by
$h^{(0)} (x) =x$ and iteratively
\begin{equation}\label{DCNN}
h^{(j)} (x) = \sigma \left(T^{(j)} h^{(j-1)}(x)  - b^{(j)}\right), \qquad j=1, 2, \ldots, J,
\end{equation}
where $T^{(j)} :=T^{w^{(j)}}=\left[w^{(j)}_{i-k}\right]_{1 \leq i \leq d_{j-1} +s, 1\leq k\leq d_{j-1}}$ is a $(d_{j-1} +s) \times d_{j-1}$ convolutional matrix.
The hypothesis space generated by this DCNN is given by
\begin{equation}\label{hypothesis}
{\mathcal H}_{J, s} =  \hbox{span}\left\{c\cdot  h^{(J)} (x)  + a: c\in\RR^{d_{J}},  a\in \RR, {\bf w}, {\bf b}\right\}.
\end{equation}
\end{definition}

We often take the bias vector $b^{(j)}$ of the so-called ``identical-in-middle" form
\begin{equation}\label{baisform}
b^{(j)}=[b_1, \cdots, b_{s-1}, \underbrace{b_s, \cdots, b_s}_{d_j - 2s + 2}, b_{d_j-s+2}, \cdots, b_{d_j}]^T
\end{equation}
with $d_j-2(s-1)$ repeated entries in the middle. This special shape of the bias vector $b^{(j)}$, together with the sparsity of the convolutional matrix $T^w$, tells us that the $j$-th layer of the DCNN involves only $(s+1)+(2s-1)=3s$ free parameters.

To reduce data redundancy, we introduce a downsampling operator ${\mathcal D}_m: \RR^K \to \RR^{\lfloor K/m\rfloor}$ with a scaling parameter $m\in\NN$ 
by
\begin{equation} \label{downsampling}
{\mathcal D}_m (v) = (v_{i m})_{i=1}^{\lfloor K/m\rfloor}, \qquad v\in \RR^K,    
\end{equation}
where $\lfloor u\rfloor$ denotes the integer part of $u>0$.
In other words, the downsampling operator ${\mathcal D}_m$ only ``picks up" the $m$-th, $2m$-th, $\ldots,\lfloor K/m\rfloor m$-th entries of $v$.

\begin{definition}
A downsampled DCNN of depth $J$ with downsampling at layer $J_1 \in\{1, \ldots, J-1\}$ has widths $d_0=d$ and
\begin{equation}
d_j =\left\{\begin{array}{cc}
d_{j-1} + s, & \hbox{if} \ j\not= J_1, \\
\lfloor (d_{j-1} + s)/d\rfloor, & \hbox{if} \ j= J_1,
\end{array}\right.
\end{equation}
and function vectors $\{h^{(j)}: \RR^d \to \RR^{d_j}\}_{j=1}^J$ given by $h^{(0)} (x) =x$ and
 iteratively
 \begin{equation}\label{deepCNN}
h^{(j)} (x) = \left\{\begin{array}{cc}
\sigma \left(T^{(j)} h^{(j-1)}(x)  - b^{(j)}\right), & \hbox{if} \ j\not= J_1, \\
\sigma \left({\mathcal D}_d \left(T^{(j)} h^{(j-1)}(x)\right)  - b^{(j)}\right), & \hbox{if} \ j= J_1.
\end{array}\right.
\end{equation}
\end{definition}

In other words, the downsampling operation aims to reduce the width of a certain layer of DCNN while preserving information on data features.
The hypothesis space is defined in the same way as (\ref{hypothesis}).

In this paper, we take bias vectors $b^{(j)}$ to satisfy (\ref{baisform}) for $j=1,2,\ldots, J-1$. If no additional constraints are imposed, the number of free parameters for an output function from the hypothesis space  (including filters and, biases, coefficients) equals

\begin{equation}\label{parameternum}
\sum_{j=1}^{J-1}\left(s+1+2s-1\right) + (s+1) + d_J+ d_J+1=3s(J-1)+s+2+2d_J.
\end{equation}

DCNNs considered in this paper are based on a ``zero-padding" approach and have increasing widths. In the literature, DCNNs without zero-padding have also been introduced \cite{he2016deep, Krizhevsky2012}, and they have decreasing widths, leading to limited approximation abilities and the necessity of channels for learning. Moreover, DCNNs induced by group convolutions were studied with nice approximation properties presented in \cite{Petersen2020}.

\subsection{Data and Regression Function}
\label{sec:data}
Consider a training sample $D:=\{z^i=(x^i, y^i)\}_{i=1}^n$  drawn independently and identically distributed from an unknown distribution $\rho$ on $Z:=\Omega \times \mathcal{Y}$.
Throughout this paper, we assume that $\Omega$ is a closed bounded subset of $\RR^d$ and $\mathcal{Y} =[-M, M]$ for some $M \geq 1$.

The regression function $f_\rho$ is given by conditional means $f_\rho (x) = {\mathbb E}[y|x] =\int_{\mathcal{Y}} y d\rho(y|x)$ of the conditional distributions $\rho(\cdot|x)$ at $x\in\Omega$. Since $|y|\leq M$ almost surely,
we have $|f_\rho(x)|=|\mathbb{E}[y|x]| \leq M$. So, it is natural to project function values onto the interval $[-M, M]$ and define the truncation operator $\pi_M$ for any real-valued function $f$ by
\begin{equation}
\pi_Mf (x) =
\begin{cases}
f(x), &  \text{if } |f(x)|\leq M,\\
M, & \text{if } f(x)>M,\\
-M, & \text{if } f(x)<-M.
\end{cases}
\end{equation}

Moreover, we denote by $\mathcal{H}_{int,J,s}$ the set of all output functions $f\in {\mathcal H}_{J, s}$ from the hypothesis space ${\mathcal H}_{J, s}$ (\ref{hypothesis}) satisfying the following interpolation condition
\begin{equation} \label{interpolation}
f(x^i)=y^i, \qquad i=1,\ldots, n.
\end{equation} 
When downsampling is used in the DCNN, we use the same notation to denote the set of interpolating output functions generated from the downsampled DCNN. 

\section{Main Results} \label{results}
In this section, we state our main results.
The corresponding theorems will be proved in Sections \ref{Proof1} and \ref{Proof2}.

Here, we introduce our first result. Our first result shows that a good generalization ability of any given non-interpolating DCNN (``teacher DCNN") can be maintained by some interpolating output function $f\in \mathcal{H}_{int, J,s }$ of a ``student DCNN" obtained by adding well-defined layers to the given DCNN. 
The learning and approximation ability is measured in regression by the $L_2$ norm $\|f\|_2 :=\left\{\int_\Omega |f(x)|^2 d \rho_\Omega\right\}^{1/2}$ for $f\in L^2_{\rho_\Omega}$ with respect to the marginal distribution $\rho_\Omega$ of $\rho$ on $\Omega$.

\begin{theorem}\label{MainResult}
Let $2\leq d\in \NN$ and $S\leq d/2$.
Assume that $\rho_\Omega$ has no positive mass at any point $x\in\Omega$.
Suppose that the hypothesis space ${\mathcal H}_{J_2, S}$ defined by (\ref{hypothesis}) generated by a DCNN of depth $J_2 \in\NN$ with filter length $S$ can approximate the regression function $f_\rho$ within accuracy $\{E_{J_2}>0\}_{J_2\in\NN}$ as follows
\begin{equation}\label{pre-error}
\inf_{f\in {\mathcal H}_{J_2,S}} \left\|\pi_M f - f_\rho\right\|_2 < E_{J_2},
\end{equation}
then there exists a downsampled DCNN of depth $J_1 + J_2 + J_3$ with filter length $s=2S$, downsampling at the $J_1$-th layer where $J_1 = \lceil \frac{2d^2}{s-1}\rceil$, and $J_3 = \left\lceil \frac{(N-1) d_{J_{1}+J_2}}{s}\right\rceil$ for some odd $N \geq 3n$,  such that
\begin{equation}\label{CNNerror}
\inf_{f\in {\mathcal H}_{int, J_1+J_2+J_3, S}} \left\|\pi_M f - f_\rho\right\|_2< E_{J_2}.
\end{equation}
The additional number of free parameters of an output function from the deepened DCNN equals
$$
d_{J_1+J_2+J_3}+
J_1(s+2)+3.
$$
\end{theorem}

More specifically, suppose we are given a DCNN of depth $J_2$, which approximates $f_\rho$ sufficiently well.
Theorem \ref{MainResult} states that, by adding $J_1 + J_3$ well-defined convolutional layers to this given DCNN, we obtain some$f\in  \mathcal{H}_{int, J_1+J_2+J_3, S}$ which interpolates the data and, at the same time, possesses the same generalization error bound
as the given DCNN.


The main tool of proving Theorem \ref{MainResult} is a network deepening scheme, which adds well-defined layers to a given DCNN such that the deepened student DCNN interpolates the input data.
The detailed proof is given in Section \ref{Proof1}.

To verify the benign overfitting of DCNNs, it is necessary to show that an interpolating DCNN can achieve a good learning rate. The phrase ``learning rate" here refers to the convergence rate of the excess generalization error (excess error), which will be defined shortly.
We now turn our attention to finding learning rates of a DCNN of depth $J$ and filter length $s$.
It will act as a teacher DCNN later in Theorem \ref{MainResult3}.
Such a DCNN generates the hypothesis space $\mathcal{H}_{J,s}$ defined in (\ref{hypothesis}).
We are interested in a global minimum of the following empirical risk minimization (ERM) problem
\begin{equation} \label{ERM}
f_{D,J,s}:= \text{arg min }_{f\in \mathcal{H}_{J,s}}\ \varepsilon_D(f),
\end{equation}
    where $\varepsilon_D(f)= \frac{1}{n}\Sigma_{i=1}^ n (f(x^i)-y^i)^2$ is the empirical risk of function $f \in C(\Omega)$ , the space of continuous functions on $\Omega$ with norm $\|f\|_{C(\Omega)} = \sup_{x\in \Omega} |f(x)|$.

To analyze the performance of learning algorithms for regression, we consider the generalization error defined by
$$\varepsilon(f) = \int_{Z} (f(x)-y)^2 \,d\rho.$$
It is minimized by the regression function $f_\rho$, and
the excess generalization error (excess error) equals the error norm square
\begin{equation}\label{excesserror}
\varepsilon(f) - \varepsilon(f_\rho) =\|f-f_\rho\|_2^2
\end{equation}
and can be used in regression analysis \cite{Gyorfi2002}.

Here, we present our second result. Our second result establishes the learning rates of DCNNs in general.
Remarkably, we do not give any restrictions on norms of the filters and bias vectors or variable structures of the regression function $f_\rho$.
The Sobolev space $W_2^{r} (\Omega)$ with an integer index $r$ consists of
restrictions to $\Omega$ of $F$ from the Sobolev space $W_2^{r} ({\mathbb R}^d)$ on ${\mathbb R}^d$ meaning that $F$ and all its partial derivatives up to order $r$ are squared integrable on ${\mathbb R}^d$.

\begin{theorem} \label{MainResult2}
Let $2 \leq S\leq d$, $M \geq 1$, $n \geq 3$, and $\Omega \subseteq [-1, 1]^d$. Assume $|y| \leq M$ almost surely and $f_\rho \in W_2^{r} (\Omega)$ with an integer index $r>2 + d/2$. Define $f_{D,J,S}$ by the ERM scheme (\ref{ERM}) with the DCNN of depth $J$ stated in Definition \ref{defDCNN} with ``identical-in-middle" bias vectors satisfying (\ref{baisform}) for layers $1, 2, \ldots J-1$.
The number of free parameters of this DCNN is $P_{D,J,S} \leq 5dJ+2$.
For any $0 <\delta <1$, we have with probability at least $1- \delta$,
\begin{equation}\label{boundJ}
\left\|\pi_M f_{D,J,S}- f_\rho\right\|_2^2 \leq C_{M, r, f_\rho} d (\log d)\left(1 + \frac{\log\left(2/\delta\right)}{\sqrt{n}}\right)\left(\frac{(\log n) J^2 (\log J)}{n} + \frac{1}{\sqrt{n}}+\frac{\log J}{J}\right)
\end{equation}
where $C_{M, r, f_\rho}$ is a constant independent of $n, \delta$ or $d$.
\\
Specifically, when $J =\lceil n^\alpha\rceil$ with any $0<\alpha<1/2$, the number of free parameter is $P_{D,\lceil n^\alpha\rceil,S} \leq 5d\lceil n^\alpha\rceil+2$.  With probability at least $1- \delta$, we have
\begin{equation}\label{boundwithoutJ}
\left\|\pi_M f_{D,J,S}- f_\rho\right\|_2^2 \leq \max\left\{8M^2,6C_{M, r, f_\rho}\right\} d (\log d) \left(1 + \frac{\log\left(2/\delta\right)}{\sqrt{n}}\right) \frac{(\log n)^2}{n^{\min\{1-2\alpha, \alpha\}}}.
\end{equation}
\end{theorem}

Theorem \ref{MainResult2} establishes learning rates of DCNNs explicitly in terms of the dimension of the input data $d$ and size $n$ of the training sample.
It states that when $J=\lceil n^\alpha\rceil$ for any $0<\alpha<1/2$, the DCNN output function $\pi_M f_{D,J,S}$ converges to the regression function $f_\rho$ with high probability.
In particular, with the choice of $J =\lceil n^{\frac{1}{3}}\rceil$, the rate of  convergence is of order $\mathcal{O}((\log n)^2n^{-1/3})$.
To our best knowledge, this is the first result presenting learning rates of DCNNs for learning a general function without any variable structure assumption. It differs from the existing learning rates of DCNNs for learning functions with variable structures such as additive ridge functions \cite{fang2022optimal} and radial functions \cite{mao2021theory}. This is also the first learning rate of DCNNs without parameter restrictions presented in the literature. The proof of Theorem \ref{MainResult2} is given in Section \ref{Proof2}.

Next, we present our third result. According to Theorem \ref{MainResult2}, for any $0<\alpha<1/2$, an underparameterized, non-interpolating DCNN with $\mathcal{O}(n^\alpha)$ free parameters converges to the regression function with a learning rate of order $\mathcal{O}((\log n)^2n^{-1/3})$. 
In this paper, we refer to ``underparameterized neural networks" \cite{Belkin2019} as networks that have the number of trainable free parameters of order $o(n)$.
Next, by applying the results in Theorem \ref{MainResult}, we can deepen this underparameterized DCNN to an interpolating, overparameterized one. The following theorem  suggests that this overparameterized DCNN with $\mathcal{O}(n^{1+\alpha})$ free parameters can not only interpolate any input data, but also achieve the same learning rate as the underparameterized DCNN.

\begin{theorem}\label{MainResult3}
Under the assumption of Theorem \ref{MainResult2} and that $\rho_\Omega$ has no positive mass at any point $x\in\Omega$,
for any $0<\alpha<1/2$,
there exists a downsampled DCNN of depth $J_1 + J_2 + J_3$ with even filter  length $s$ satisfying $4 \leq  s \leq d$, downsampling at the $J_1$-th layer, where $J_1 = \lceil \frac{2d^2}{s-1}\rceil$, $J_2=\lceil n^{\alpha}\rceil$,and $J_3 = \left\lceil \frac{4n \left(d_{J_{1}}+s\lceil n^{\alpha}\rceil\right)}{s}\right\rceil$,  such that for any $0<\delta<1$, with probability $1-\delta$, we have
\begin{equation}
\inf_{f\in {\mathcal H}_{int, J_1+ J_2+J_3,s} }\left\|\pi_M f - f_\rho\right\|_2^2 < \max\left\{8M^2+1 ,6C_{M, r, f_\rho }+1\right\} d (\log d) \left(1 + \frac{\log\left(2/\delta\right)}{\sqrt{n}}\right) \frac{(\log n)^2}{n^{\min\{1-2\alpha, \alpha\}}}.
\end{equation}
The number of free parameters of this DCNN is of order  $\mathcal{O}(n^{1+\alpha})$.
\end{theorem}

Theorem \ref{MainResult3} tells us that there exists an interpolating DCNN that generalizes well. This result provides theoretical support for the possible existence of benign overfitting in the DCNN setting. 
At this point, we are unable to  derive learning rates of DCNN with $J=\lceil n^\alpha\rceil$ layers for $1/2 \leq  \alpha < 1$.
The challenge is mainly due to an upper bound of pseudo-dimension  stated in Theorem 4 below (an extension to a previous result in \cite{Bartlett2019}), which does not cover the case $1/2 \leq  \alpha < 1$. It would be interesting to develop new approaches to estimate the pseudo-dimension of DCNNs, which in turn estimates the covering number.
Furthermore, it is worth noting that  the learning rates we obtained (for both underparameterized and overparameterized DCNNs) do not achieve the minimax rate of convergence for least squares regression \cite{Schmidt-Hieber}.
Whether one can obtain a minimax rate of convergence of DCNNs remains an open problem.

\section{Achieving Interpolation Condition by Deepening DCNNs} \label{Proof1}

In this section, we present a network deepening scheme for proving Theorem \ref{MainResult}. 
Here, we first give an outline of the proof.

Suppose we are given a non-interpolating DCNN that outputs a function $f^*$. Theorem 1 tells us that we can construct a larger DCNN that interpolates any data while maintaining the generalization ability of $f^*$. The proof is carried out by means of an interpolator of the form 
\begin{equation}\label{interpformBrief}
f(x) =f^{*}(x) +  \sum_{\ell=1}^n \left(y^\ell - f^{*}(x^\ell)\right) \phi_\ell (x) \quad \hbox{with} \quad \phi_\ell (x)=\phi\left(\eta^{*} \xi \cdot (x  - x^\ell)\right),
\end{equation}
where $\eta^{*} \in \{1, -1\}$ is a sign number, $\xi\in\RR^d$ is a nonzero vector, and $\phi =\phi^{(\epsilon)}: \RR \to \RR$ given with some $\epsilon>0$ by
\begin{align*}
  \phi(u) &= \frac{1}{\epsilon} \left\{\sigma\left(u+ \epsilon\right) - \sigma\left(u\right)\right) - \left(\sigma\left(u\right) - \sigma\left(u- \epsilon\right)\right\} \\
  &= \begin{cases}
     1+ \frac{1}{\epsilon}u & \text{if } -\epsilon \leq u <0,\\
     1-\frac{1}{\epsilon}u & \text{if } 0 \leq u \leq \epsilon,\\
     0 & \text{if } |u|>\epsilon.
    \end{cases}  
\end{align*}
We can see that $\phi$ is a hat function vanishing outside $[-\epsilon, \epsilon]$, equal to $1$ at $0$, and linear on $[-\epsilon, 0]$ and on $[0, \epsilon]$. 
When $\epsilon$ is small enough, the basis function $\phi_\ell$ in (\ref{interpformBrief}) 
takes the form \begin{equation*}
    \phi_\ell(x^j)=\begin{cases}
                 1 & \text{if } j=\ell,\\
                 0 & \text{if } j \neq \ell,
    \end{cases}  
\end{equation*}
leading to the desired interpolation property (that is, $f(x^i)=y^i$ for all $i$) while maintaining the generalization abilities of $f^{*}$. 
Our goal is to construct a DCNN that produces this function $f$. The construction of such a DCNN can be divided into three steps.

First, we use a group of convolutional layers to realize linear features $\xi \cdot x$ in the basis functions $\phi_\ell$ for an arbitrary $\xi \in \RR^d$ (Subsection \ref{sec:construct-linear-feature}).
Then, we introduce a network deepening scheme for any given DCNN (Subsection \ref{sec:deepening}).
It doubles the widths of the given DCNN, maintains its learning rate, and at the same time embeds the linear features $\xi \cdot x$. This is the key novel idea in our deepening scheme.  
Lastly, we use ridge functions to construct DCNN interpolators (Subsections \ref{sec:construct-cnn-for-linear} and \ref{sec:achieve-interpolate}). The overview of this network construction is illustrated in Figure 1.

\begin{figure} 
\centering
\includegraphics[width=0.9\textwidth]{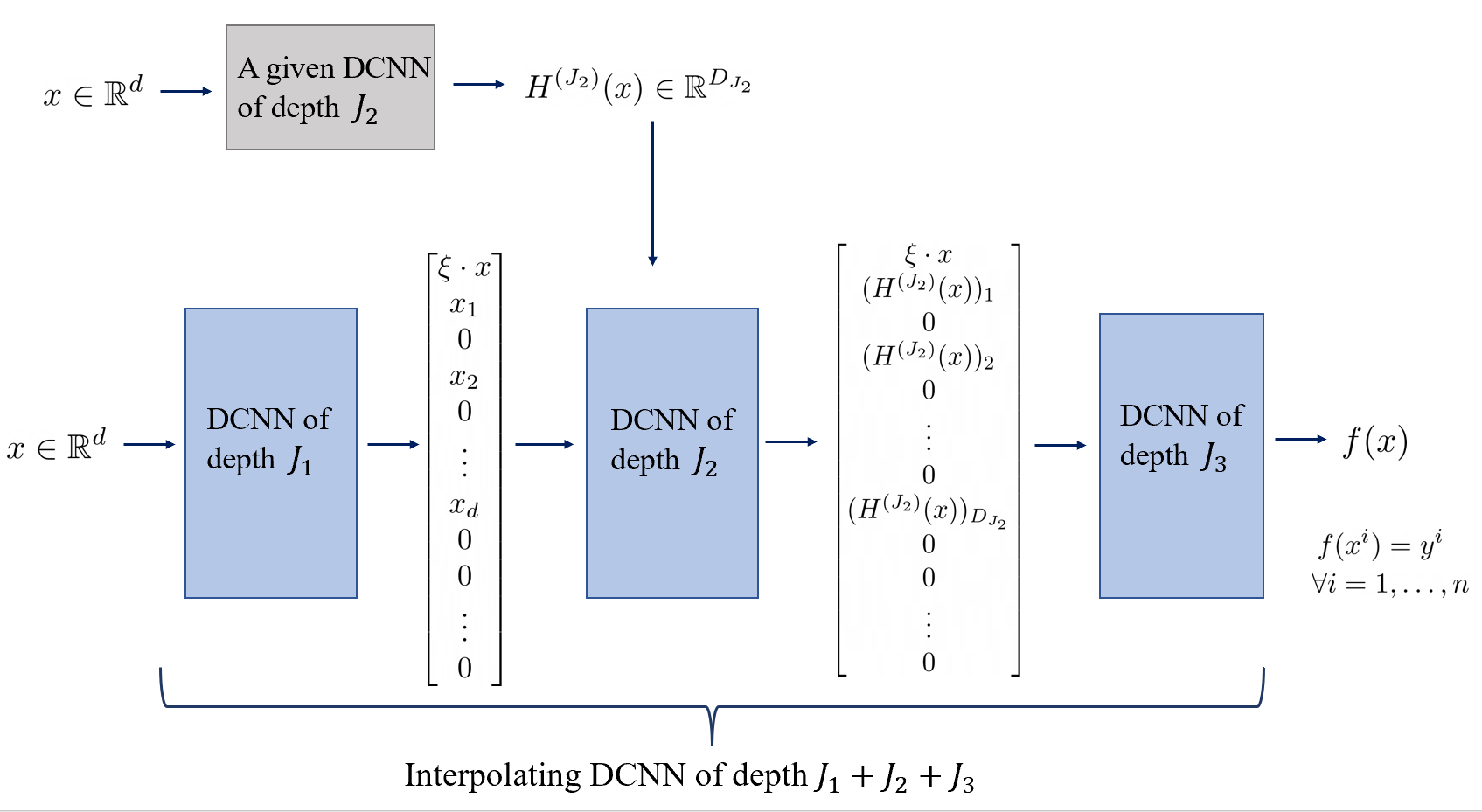}
\caption{Given an input $x\in \RR^d$ and a given non-interpolating DCNN of depth $J_2$, the figure illustrates the process of constructing an interpolating DCNN of depth $J_1+J_2+J_3$.}
\end{figure}

Before we get into these three steps, we first introduce a proposition. Suppose we have any  given sequence $W$ and network input $U(x)$, where $U\in (C(\Omega))^K$ with norm $\|U\|_\infty = \sup_{x\in \Omega} \|U(x)\|_{\ell^\infty (\RR^K)}$. This proposition suggests that we can construct a DCNN that outputs $T^W U(x)$, where $T^W$ is a convolutional matrix induced by $W$. Later in constructing linear features (the first step) and ridge functions (the third step), we will apply this result to produce some special vectors truncated from a sequence $W$. 

Denote ${\bf 1}_{K} = (1, \ldots, 1) \in \RR^{K}$. 
Recall the  ``identical-in-middle" bias vector introduced in (\ref{baisform}): \begin{equation*}
b^{(j)}=[b_1, \cdots, b_{s-1}, \underbrace{b_s, \cdots, b_s}_{d_j - 2s + 2}, b_{d_j-s+1}, \cdots, b_{d_j}]^T.
\end{equation*}

\begin{proposition}\label{generalCNN}
If $W=(W_k)_{k=-\infty}^{\infty}$ is a sequence supported in $\{0, \ldots, {\mathcal V}\}$, $K\in\NN$ is the input dimension, and $2\leq s \leq K$, 
then there exist filters $\{w^{(j)}\}_{j=1}^{J^{*}}$ each supported in $\{0, \ldots, s\}$ with $J^{*} = \lceil \frac{{\mathcal V}}{s-1}\rceil$ 
such that the convolutional factorization
$W  = w^{(J^{*})}{*}w^{(J^{*}-1)}{*}\cdots {*}w^{(2)}{*}w^{(1)}$ holds. The filter $w^{(i)}$ 
induces a $(K + i s) \times (K + (i-1)s)$ convolutional matrix $T^{(i)}$ given by (\ref{convolution}). 

Consider a DCNN $\left\{\hat{h}^{(j)}(x): \RR^K \to \RR^{{K +js}}\right\}_{j=0}^{J^{*}}$ 
satisfying $\hat{h}^{(j)} (x) = \sigma \left(T^{(j)} \hat{h}^{(j-1)}(x)  - b^{(j)}\right)$. 
If the input layer 
$\hat{h}^{(0)}$ is given by $U(x) + \vec C$ with $\vec C \in \RR^{K}$ and $\|U\|_\infty\leq {\mathcal B}^{(0)}$ for some ${\mathcal B}^{(0)}>0$, 
and the bias vectors are given by $b^{(1)} =T^{(1)} \vec C - \|w^{(1)}\|_1 {\mathcal B}^{(0)} {\bf 1}_{K+s}$, and
\begin{equation} \label{biasbjD}
b^{(j)} =
\left\{\prod_{i=1}^{j-1} \|w^{(i)}\|_1\right\} {\mathcal B}^{(0)} T^{(j)} {\bf 1}_{K + (j-1)s} - \left\{\prod_{i=1}^j \|w^{(i)}\|_1\right\} {\mathcal B}^{(0)} {\bf 1}_{K + js}, \qquad j=2, \ldots, J^{*}-1,   \end{equation}
then $b^{(j)}$ satisfy the  ``identical-in-middle" condition (\ref{baisform}) for $j=2, \ldots, J^{*}-1$, the following expressions hold
\begin{equation}\label{iterationhjD}
\hat{h}^{(j)} (x) = T^{(j)}T^{(j-1)}\cdots T^{(1)} U(x) + \left\{\prod_{i=1}^j \|w^{(i)}\|_1\right\} {\mathcal B}^{(0)} {\bf 1}_{K + js}, \quad j=1, \ldots, J^{*} -1, 
 \end{equation}
and 
\begin{equation}\label{iterationhJ}
T^{(J^{*})} \hat{h}^{(J^{*} -1)} (x)  =T^W U(x) + \left\{\prod_{i=1}^{J^{*}-1} \|w^{(i)}\|_1\right\} {\mathcal B}^{(0)} T^{(J^{*})}\left({\bf 1}_{K + (J^{*}-1)s}\right),
 \end{equation}
 where $T^W = \left[W_{\ell -k}\right]_{\ell, k}\in \RR^{(K+J^*s)\times K}$. 
Moreover, if $\vec C ={\mathcal C} {\bf 1}_K$ for some 
${\mathcal C} \geq 0$, then (\ref{baisform}) is also satisfied for $j=1$. 
\end{proposition}

The expression (\ref{iterationhJ}) will be applied in the first and third steps of our construction, with different choices of $W$. 

\remark{The convolutional matrix $T^W$ is a $(K + J^{*} s)\times K$ Toeplitz-type matrix induced by $W$ and is given by 
\begin{equation}\label{matrixTW}
T^{W} = \left[\begin{array}{llll}
W_0 & 0 & \cdots  & 0 \\
W_1 & W_0 & \ddots  & 0  \\
\vdots & \ddots & \ddots & \vdots \\
W_{K-1} & \cdots &  W_1 & W_0 \\
W_{K} & \ddots &  \ddots & W_1  \\
\vdots & \ddots &  \ddots & \vdots  \\
 W_{2K-1} & W_{2K-2} & \cdots & W_K\\
\vdots & \ddots &  \ddots & \vdots  \\
W_{J^{*} s} & \cdots & \ddots & \vdots \\
0 & W_{J^{*} s} & \cdots & \vdots \\
\vdots & \ddots & \ddots & \vdots  \\
0 & \cdots & 0 & W_{J^{*} s}
\end{array}\right].
\end{equation}}
If we apply the downsampling operator  ${\mathcal D}_K: \RR^{K+J^*s} \to \RR^{\lfloor (K+J^*s)/K\rfloor}$ in (\ref{downsampling}) to the last layer of DCNN proposed in Proposition \ref{generalCNN}, we get 
\begin{equation} \label{downsampledTW}
    {\mathcal D}_K \left(T^{(J^{*})} \hat{h}^{(J^{*} -1)} (x)\right) =  \left[\begin{array}{llll}
W_{K-1} & \cdots &  W_1 & W_0 \\
W_{2K-1} & W_{2K-2} & \cdots & W_K\\
W_{3K-1} & W_{2K-2} & \cdots & W_{2K}\\
\vdots & \ddots &  \ddots & \vdots  \\
W_{K + K\lfloor J^{*} s/K \rfloor} -1 & \cdots & \ddots & W_{K\lfloor J^{*} s/K \rfloor} \\
\end{array}\right]U(x) + b,
\end{equation}
where $b$ is some constant vector. We see that all matrix entries in (\ref{downsampledTW}) are potentially different. Later, we use this important observation to generate specific features using DCNNs. 

Before we prove Proposition \ref{generalCNN}, we first introduce two supporting Lemmas. 

Lemma \ref{filterdecom} was stated as Theorem 3 in \cite{zhou2020universality}. It suggests that any sequence $W$ can be factorized into convolutions of some smaller sequences. 
The filters $\{w^{(j)}\}_{j=1}^{J^{*}}$ in Proposition \ref{generalCNN}  are generated according to this result.
\begin{lemma}\label{filterdecom}
If $W=(W_k)_{k=-\infty}^{\infty}$ is supported in $\{0, \ldots, {\mathcal V}\}$, then there exist filters $\{w^{(j)}\}_{j=1}^{p}$ each supported in $\{0, \ldots, s\}$ with $p \leq \lceil \frac{{\mathcal V}}{s-1}\rceil$ satisfying a convolutional factorization
$$
W  = w^{(p)}{*}w^{(p-1)}{*}\cdots {*}w^{(2)}{*}w^{(1)}.
$$
\end{lemma}

A nice property stated in the following lemma (Lemma \ref{matrixproducts}) is that products of convolutional matrices $\prod_{i=1}^j T^{(i)}$ 
are still convolutional matrices but induced by the convoluted sequence of larger support. The proof of Lemma \ref{matrixproducts} is given in Appendix A.1.

\begin{lemma}\label{matrixproducts}
Let $\{w^{(j)}\}_{j=1}^{J^{*}}$ be filters supported in $\{0, \ldots, s\}$, then for $j=1, \ldots, J^{*}$, we have
\begin{equation}\label{filterI}
\prod_{i=1}^j T^{(i)} = T^{{\bf w}, j},
\end{equation}
where $\prod_{i=1}^j T^{(i)}= T^{(j)}T^{(j-1)}\cdots T^{(1)}$, and $T^{{\bf w}, j} :=\left[W^{(j)}_{\ell -k}\right]_{\ell=1, \ldots, K + j s, k=1, \ldots, K}$ is the $(K+js) \times K$ convolutional matrix induced by the convoluted sequence $W^{(j)} =w^{(j)}{*}w^{(j-1)}{*}\cdots {*}w^{(1)}$ supported in $\{0, \ldots, j s\}$.
\end{lemma}

Now we are in a position to prove Proposition \ref{generalCNN}. 

\begin{proof}[Proof of Proposition \ref{generalCNN}] 
We apply Lemma \ref{filterdecom} and find filters $\{w^{(j)}\}_{j=1}^{p}$ with $p \leq J^*= \lceil \frac{{\mathcal V}}{s-1}\rceil$ satisfying 
$
W  = w^{(p)}{*}w^{(p-1)}{*}\cdots {*}w^{(2)}{*}w^{(1)}.
$
When $p<J^*$, we choose each filter $w^{(j)}$ with $j=p+1, \ldots, J^{*}$ to be the delta sequence which 
takes the value $1$ at $0$ and the value $0$ at any other integers (that is, $w^{(j)}_0 =1$ and $w^{(j)}_i = 0$ for $i\not=0$). Then the desired convolutional factorization follows. 

Note that the Toeplitz-type matrix 
$T^{(j)}$ induced by the filter $w^{(j)}$ has the form 
\begin{equation}\label{matrixTj}
T^{(j)} = \left[\begin{array}{llllllllll}
w^{(j)}_0 & 0 & 0 & \cdots & \cdots & 0  &0 \\
\vdots  & \ddots& \ddots & \ddots & \ddots & \ddots  & \vdots \\
        w_s^{(j)} &  \cdots &  w_0^{(j)} &  0 &\cdots & 0  & 0 \\
0 & w_s^{(j)} & \cdots & \ w_0^{(j)} & 0 &\cdots & 0 \\
\vdots & \ddots & \ddots  & \ddots&\ddots & \ddots & \vdots  \\
\vdots & \ddots & \ddots  & \ddots&\ddots & \ddots & \vdots  \\
0 & \cdots & \cdots  & 0 &  w_{s}^{(j)}&  \cdots  & w_0^{(j)} \\
\vdots & \ddots &  \ddots &  \ddots&  \ddots & \ddots &  \vdots  \\
0 & 0&\cdots &  \cdots & 0 &0& w_{s}^{(j)}
\end{array}\right] .
\end{equation} 
We made an important observation that the entry sum of each row in the middle of $T^{(j)}$ equals $\sum_{\ell =0}^s w^{(j)}_\ell$. Then, 
we see that the choice (\ref{biasbjD}) of the bias vector $b^{(j)}$ is indeed a ``identical-in-middle" bias vector satisfying 
(\ref{baisform}) for $j=2, \ldots, J^{*}-1$. It is also satisfied for $j=1$ if $\vec C ={\mathcal C} {\bf 1}_K$ for some ${\mathcal C} \geq 0$. 

Then we prove the expression (\ref{iterationhjD}) by induction. Since the row sums of the Toeplitz-type matrix $T^w$ is bounded by $\|w\|_1=\sum_{k=-\infty}^{\infty} |w_k|$, we know that
$\|T^w v\|_{\infty} \leq \|w\|_1 \|v\|_\infty$ for any $v\in\RR^K$. The case $j=1$ holds because 
$T^{(1)} \hat{h}^{(0)}(x)  - b^{(1)} =T^{(1)} U(x)  + \|w^{(1)}\|_1 {\mathcal B}^{(0)} {\bf 1}_{K+s}$ has nonnegative entries 
due to the fact that each component of $T^{(1)} U(x)$ is bounded by $\|w^{(1)}\|_1 \|U(x)\|_\infty \leq \|w^{(1)}\|_1 {\mathcal B}^{(0)}$. 

If (\ref{iterationhjD}) is valid for $j-1$, that is, 
$$\hat{h}^{(j-1)} (x) = T^{(j-1)}\cdots T^{(1)} U(x) + \left\{\prod_{i=1}^{j-1} \|w^{(i)}\|_1\right\} {\mathcal B}^{(0)} {\bf 1}_{K + (j-1)s}, $$
then 
$$T^{(j)} \hat{h}^{(j-1)}(x)  - b^{(j)} =T^{(j)}T^{(j-1)}\cdots T^{(1)} U(x) + \left\{\prod_{i=1}^j \|w^{(i)}\|_1\right\} {\mathcal B}^{(0)} {\bf 1}_{K + js} $$
has nonnegative entries since 
$$ \|T^{(j)}T^{(j-1)}\cdots T^{(1)} U(x)\|_\infty \leq \left\{\prod_{i=1}^j \|w^{(i)}\|_1\right\} {\mathcal B}^{(0)}. $$
Hence, \begin{align*}
 \hat{h}^{(j)} (x) &= \sigma \left(T^{(j)} \hat{h}^{(j-1)}(x)  - b^{(j)}\right) \\&=  \sigma \left(T^{(j)}T^{(j-1)}\cdots T^{(1)} U(x) + \left\{\prod_{i=1}^j \|w^{(i)}\|_1\right\} {\mathcal B}^{(0)} {\bf 1}_{K + js}\right)\\
 &=T^{(j)}T^{(j-1)}\cdots T^{(1)} U(x) + \left\{\prod_{i=1}^j \|w^{(i)}\|_1\right\} {\mathcal B}^{(0)} {\bf 1}_{K + js} 
\end{align*}
This completes the induction procedure 
and verifies (\ref{iterationhjD}). The expression for $T^{(J^{*})}\hat{h}^{(J^{*}-1)} (x)$ follows easily. The proof of Proposition \ref{generalCNN} is complete. 
\end{proof}

\subsection{The First Step: Realising linear functions by DCNNs}
\label{sec:construct-linear-feature} 

The first step described in this subsection shows how to realize a linear feature $\xi \cdot x$, i.e., the inner product of vectors $\xi$ and $x$, by the first $J_1$ layers of a DCNN. 
In this step, we apply Proposition \ref{generalCNN} and the downsampling operation. This idea of generating linear features has been introduced for DCNNs in \cite{zhou2020universality} and for periodized DCNNs using circular weight matrices in \cite{Petersen2020}. We choose a special sequence $W$ such that the output of this step is a function vector containing  $\xi \cdot x$ and $x_i$.

By choosing $K=d$, $[W_{d-1} \ \cdots \ W_1 \ W_0]$ to be the vector $\xi$ and $[W_{i d + d-1} \ \cdots \ W_{i d +1} \ W_{i d}]$ to be the $i$-th basis vector $e_i$ for $i=1, \ldots, d$, we can make the function vector on the right-hand side of (\ref{downsampledTW}) to have components $\xi \cdot x$ and $x_i$.

\begin{proposition}\label{realizinglinearfeatures}
Let $2 \leq s\leq d$, $J_1 = \lceil \frac{2d^2-1}{s-1}\rceil$, $\xi \in \RR^d$, $\Omega$ be a closed bounded subset of $\RR^d$.Take a constant $B^{(0)} \geq \max_{x\in \Omega} \|x\|_\infty$. Then there exist filters $\{w^{(j)}\}_{j=1}^{J_1}$ each supported in $\{0, \ldots, s\}$ such that with $b^{(1)} =- \|w^{(1)}\|_1 B^{(0)} {\bf 1}_{d+s}$, $\{b^{(j)}\}_{j=2}^{J_1 -1}$ given by (\ref{biasbjD}) with $K=d, J^{*} = J_1, {\mathcal B}^{(0)} = B^{(0)}$ and 
$$b^{(J_1)} = \left\{\prod_{i=1}^{J_1 -1} \|w^{(i)}\|_1\right\} B^{(0)} T^{(J_1)}\left({\bf 1}_{d + (J_1 -1)s}\right) + \left\{\prod_{i=1}^{J_1} \|w^{(i)}\|_1\right\} B^{(0)} {\bf 1}_{d + J_1 s}, $$
the first $J_1$ layers of the DCNN $\left\{h^{(j)}(x): \RR^d \to \RR^{{d_j}}\right\}_{j=0}^{J_1}$ defined by (\ref{deepCNN}) with $J_0 =J_1$ produces
$$ h^{(J_1)} (x)= \left[
\xi \cdot x,
x_1,
0,
x_2,
0,
x_3,
0,
\cdots,
x_d,
\underbrace{
0,0,
\cdots,
0}_{\lfloor J_1 s/d\rfloor-2d+1}
\right]^T + B^{(J_1)} {\bf 1}_{1 + \lfloor J_1 s/d\rfloor},\qquad x\in \Omega.$$
Here $\{w^{(j)}, \{b^{(j)}\}_{j=1}^{J_1}$ and the constant $B^{(J_1)} :=\left\{\prod_{i=1}^{J_1} \|w^{(i)}\|_1\right\} B^{(0)}$ depend on $\xi, d, s, \Omega$.  The bias vectors $\{b^{(j)}\}_{j=1}^{J_1}$ are ``identical-in-middle" satisfying (\ref{baisform}).
The number of free parameters of the first $J_1$ layers equals $J_1(s+2)+1$.
\end{proposition}

\begin{proof}
Denote the standard basis of $\RR^d$ by $\{e_i\}_{i=1}^d$ and the zero vector in $\RR^d$ by $0_d$. Note that $\{e_i\}_{i=1}^d$ and $0_d$ are column vectors.
Let $W$ be a sequence supported on $\{0, 1, \ldots, 2d^2-1\}$ given by
 \begin{equation} \label{Wxi}
 \left[W_{2d^2-1}, \ \cdots \ ,W_0\right] = \left[e_d^T; 0_d^T;  e_{d-1}^T;  0_d^T; \ \cdots ; e_2^T;  0_d^T;  e_{1}^T;  \xi^T\right].     
 \end{equation}
Then by Proposition \ref{generalCNN} with ${\mathcal V} =2d^2-1$, there exist filters $\{w^{(j)}\}_{j=1}^{J_1}$ each supported in $\{0, \ldots, s\}$ with $J_1 = \lceil \frac{2d^2-1}{s-1}\rceil$ satisfying a convolutional factorization $W  = w^{(J_1)}{*}w^{(J_1-1)}{*}\cdots {*}w^{(2)}{*}w^{(1)}.$  Each filters $w^{(j)}$ induces a $(K+js) \times (K+(j-1)s)$ convolutional  matrix $T^{(j)}$. 

Take $K=d, U(x)=x, \vec C =0, J^* = J_1$. Then the conditions of Proposition \ref{generalCNN} are satisfied, and the chosen bias vectors satisfy (\ref{baisform}). By the expression (\ref{iterationhJ}) in Proposition \ref{generalCNN}, the resulting DCNN $\left\{h^{(j)}(x): \RR^d \to \RR^{{d_j}}\right\}_{j=0}^{J_1}$ defined by (\ref{deepCNN}) satisfies $$
T^{(J_1)} h^{(J_1 -1)} (x)  =T^W x + \left\{\prod_{i=1}^{J_1-1} \|w^{(i)}\|_1\right\} B^{(0)} T^{(J_1)}\left({\bf 1}_{d + (J_1 -1)s}\right). $$
 We apply the downsampling operator ${\mathcal D}_d$ in the $J_1$-th layer and obtain 
$$h^{(J_1)} (x) = {\mathcal D}_d\left(T^{(J_1)} h^{(J_1-1)} (x) - b^{(J_1)}\right) = {\mathcal D}_d \left(T^{W} x + B^{(J_1)}{\bf 1}_{d+ J_1s}\right).$$
Note that $T^W = \left[W_{\ell -k}\right]_{\ell=1, \ldots, d + J_1 s, k=1, \ldots, d}$ is the $(d+J_1s) \times d$ convolutional matrix induced by the sequence $W$ supported on $\{0, 1, \ldots, 2d^2-1\}$ given by (\ref{Wxi}). 
As we remarked for the downsampled Toeplitz-type matrix (\ref{downsampledTW}), the first component of ${\mathcal D}_d \left(T^{W} x\right)$ is the product of  the $d$-th row $\left[W_{d-1} \ \cdots \ W_0\right] =\xi^T$ of $T^W$ with $x$ and identical to  the linear feature $\xi^T x = \xi \cdot x$. 
The $2i$-th component of ${\mathcal D}_d \left(T^{W} x\right)$ for $i=1, \ldots, d$ is the product of the $2 i d$-th row $\left[W_{2 i d-1},\ \cdots \ , W_{(2 i -1)d}\right] = e_i^T$ of $T^W$ with $x$ and is exactly $e_i^T x = x_i$, while the $(2i-1)$-th component is given by the $(2i-1)d$-th row which is a zero row and thereby vanishes. The $(id)$-th row of $T^W$ with $i \geq 2d +1$ is $\left[W_{i d-1}, \ \cdots \ ,W_{(i -1)d}\right]$ which is also the zero vector. It follows that 
\begin{align*}
h^{(J_1)} (x) &={\mathcal D}_d (T^W x) + {\mathcal D}_d( B^{(J_1)} {\bf 1}_{d + J_1 s})\\&= 
\left[
\xi \cdot x,
x_1,
0,
x_2,
0,
\cdots,
x_d,
\underbrace{
0,0,
\cdots,
0}_{\lfloor J_1 s/d\rfloor-2d+1}
\right]^T + B^{(J_1)}{\bf 1}_{1 + \lfloor J_1 s/d\rfloor}.
\end{align*} 
This yields our desired expression of $h^{(J_1)} (x)$. 

The number of free parameters from filters equals $J_1(s+1)$, whereas the number of free parameters from biases equals $J_1+1$. The proof is complete.
\end{proof}
Note that $J_1 s/d \geq \frac{2d^2-1}{s-1} \frac{s}{d} >2d-1$.
Hence the width of $h^{(J_1)} (x)$ equals $d_{J_1} =1 + \lfloor J_1 s/d\rfloor \geq 2 d$.
Also, we have $\|\xi\|_1 \leq \|W\|_1 \leq \prod_{i=1}^{J_1} \|w^{(i)}\|_1 $.
So we have $\|\xi\cdot x\|_\infty \leq B^{(J_1)}$.

\subsection{The Second Step: Deepening a given DCNN}
\label{sec:deepening}
Now we continue to the second step of our construction. In this subsection, we introduce a novel network deepening scheme. Applying this scheme to a given DCNN, we can construct a deeper and wider DCNN that embeds the linear feature $\xi \cdot x$ while preserving the output of the given DCNN.

Suppose we are given a DCNN (as a teacher net) $\left\{H^{(j)}(x): \RR^d \to \RR^{{D_j}}\right\}_{j=0}^{J_2}$ with input $H^{(0)}(x)=x \in \Omega$, depth $J_2 \in \NN$ and filter length $S \leq d/2$  satisfying $D_j = d + j S$. Each layer of this DCNN has the form 
\begin{equation}\label{deepCNNH}
H^{(j)} (x) = \sigma \left({\overset{\circ}{T}}^{(j)} H^{(j-1)}(x)  - \overset{\circ}{b}^{(j)}\right), \qquad j=1, 2, \ldots, J_2,
\end{equation}
where $\left\{{\overset{\circ}{T}}^{(j)} = T^{{\overset{\circ}{w}}^{(j)}}\right\}_{j=1}^{J_2}$ are  the convolutional matrices generated by the filter sequence
${\bf \overset{\circ}{w}} =\left({\overset{\circ}{w}}^{(j)}\right)_{j=1}^{J_2}$
and  ${\bf b}=\left\{\overset{\circ}{b}^{(j)} \in \RR^{D_j}\right\}_{j=1}^{J_2}$ is a bias vector sequence.

Now we construct a student DCNN with filter length $s=2 S$ and depth $J_1+J_2$, where $J_1 = \lceil \frac{2d^2-1}{s-1}\rceil$, $J_2\in \NN$ to be determined later. The first $J_1$ layers are exactly the DCNN described in Proposition \ref{realizinglinearfeatures}.
For $j=J_1 +1, \ldots, J_1 + J_2$, we denote
$$  d_j = d_{J_1} + (j-J_1) s, \quad B^{(j)} = \prod_{i=1}^{j-J_1}\left\|{\overset{\circ}{w}}^{(i)}\right\|_1 B^{(J_1)}.$$
We define the filter $w^{(j)}$ supported in the set of even integers $\{0, 2, 4, \ldots, 2S =s\}$ by
$$ \left(w^{(j)}_{2 i}\right)_{i=0}^S =
\overset{\circ}{w}^{(j-J_1)}. $$
Taking the filter to be supported on the set of even integers and identical to that of the teacher net is our novelty. The special form of this filter gives a convolutional matrix $T^{(j)} = \left[w^{(j)}_{k-\ell}\right]$ as
\begin{equation}\label{matrixTjEven}
T^{(j)} = \left[\begin{array}{llllllllll}
{\overset{\circ}{w}}^{(j-J_1)}_0 & 0 & 0 & \cdots & \cdots   &0 \\
0 & {\overset{\circ}{w}}^{(j-J_1)}_0 & 0 & \cdots & \cdots   &0 \\
{\overset{\circ}{w}}^{(j-J_1)}_1 & 0 & {\overset{\circ}{w}}^{(j-J_1)}_0 & 0  \cdots & \cdots   &0 \\
\vdots  & \ddots& \ddots & \ddots & \ddots   & \vdots \\
{\overset{\circ}{w}}^{(j-J_1)}_{i} & 0 & {\overset{\circ}{w}}^{(j-J_1)}_{i-1} & 0 & \cdots &  \cdots\\ 
0 & {\overset{\circ}{w}}^{(j-J_1)}_{i} & 0 & {\overset{\circ}{w}}^{(j-J_1)}_{i-1} & 0 &  \cdots \\
\vdots & \ddots &  \ddots &  \ddots&  \ddots  &  \vdots  \\
\vdots & \ddots &  \ddots &  \ddots&  \ddots  &  \vdots  \\
0 & 0&\cdots &  \cdots & 0 & w_{s}^{(j-J_1)}
\end{array}\right].
\end{equation} 
It implies an essential property of our novel construction: the even entries of the student net output capture all the output entries of the teacher net.
We can see that when the even entries of $h^{(j-1)}(x)$ and $b^{(j)}$ are identical to those of $H^{(j-J_1 -1)}(x)$ and ${\overset{\circ}{b}}^{(j-J_1)}$ respectively, 
the even entries of $h^{(j)}(x)$ are kept the same as those in $H^{(j-J_1)}(x)$. The odd entries are made zero by taking large enough biases except for the first entry, which is kept  in the form $\left(\Pi_{i=1}^{j-J_1} {\overset{\circ}{w}}^{(i)}_0\right) \xi \cdot x  + B^{(j)}$, by means of the special form of the first row in (\ref{matrixTjEven}). This leads to the following proposition.
We denote $\delta_{j,k} = \begin{cases}
1, &\text{if }j=k,\\
0, &\text{if }j\neq k.\end{cases}$

\begin{proposition}\label{realizinglearningrates}
Define the CNN layers $\left\{h^{(j)}(x): \RR^d \to \RR^{{d_j}}\right\}_{j=J_1 +1}^{J_1 + J_2}$
by (\ref{DCNN}), where  
$$b^{(j)} = \delta_{j, J_1 +1} B^{(J_1)} T^{(J_1 +1)} {\bf 1}_{d_{j-1}} + \beta^{(j)}, \qquad j=J_1 +1, \ldots, J_1 + J_2$$ 
with the vector $\beta^{(j)} \in \RR^{d_j}$ given by
$$\beta^{(j)}_i = \left\{\begin{array}{ll} 
\left(1-\delta_{j, J_1}\right) {\overset{\circ}{w}}^{(j-J_1)}_0 B^{(j-1)} - B^{(j)}, & \hbox{if} \ i=1, \\
{\overset{\circ}{b}}^{(j-J_1)}_{i/2}, & \hbox{if} \ i\in \left\{2, 4, 6, \ldots, 2 D_{j-J_1}\right\}, \\
\left\|{\overset{\circ}{w}}^{(j-J_1)}\right\|_1 \left\|H^{(j-J_1-1)}\right\|_\infty + 2 B^{(j)}, & \hbox{if} \ i\not\in\left\{1, 2, 4, 6, \ldots, 2 D_{j-J_1}\right\}. 
\end{array}\right. $$
Then for $j=J_1 +1, \ldots, J_1 +J_2$, we have
\begin{eqnarray*}
&&h^{(j)} (x) = \Bigl[
\left(\Pi_{i=1}^{j-J_1} \overset{\circ}{w}^{(i)}_0\right) \xi \cdot x  + B^{(j)}, \quad
\left(H^{(j -J_1)} (x)\right)_1, 0, \
\left(H^{(j-J_1)} (x)\right)_2, \\
&& \qquad 0,\ \left(H^{(j-J_1)} (x)\right)_3, 0, \cdots \ ,
0 ,
\left(H^{(j-J_1)} (x)\right)_{D_{j-J_1}}, \
0, \
\cdots \ ,
0\Bigr]^T.
\end{eqnarray*}
No new free parameters are required for constructing $J_1+1,\ldots, J_1+J_2$-th layers.
\end{proposition}

\begin{proof} We prove by induction that for $j=J_1, J_1+1, \ldots, J_1 +J_2$, 
\begin{eqnarray}
&&h^{(j)} (x) = \Bigl[
\left(\Pi_{i=1}^{j-J_1} \overset{\circ}{w}^{(i)}_0\right) \xi \cdot x  + \left(1-\delta_{j, J_1}\right) B^{(j)}, \quad
\left(H^{(j -J_1)} (x)\right)_1, 0, \
\left(H^{(j-J_1)} (x)\right)_2,\nonumber\\
&& \qquad 0,\ \left(H^{(j-J_1)} (x)\right)_3, 0, \cdots \ ,
0,
\left(H^{(j-J_1)} (x)\right)_{D_{j-J_1}}, \
0, \
\cdots \ ,
0\Bigr]^T + \delta_{j, J_1} B^{(j)} {\bf 1}_{d_{j}}. \label{iterationhjrevise}
\end{eqnarray}
The case $j=J_1$ is trivial by Proposition \ref{realizinglinearfeatures} and the input layer $H^{(0)}(x)=x$ of the teacher net.

Suppose the claim is true for $j-1$ with $j \geq J_1 +1$. Then we see in either of the  two cases $j-1=J_1$ and $j-1>J_1$ that the first component of $h^{(j-1)} (x)$ is $\left(h^{(j-1)} (x)\right)_1 = \left(\Pi_{i=1}^{j-J_1 -1} \overset{\circ}{w}^{(i)}_0\right) \xi \cdot x  + B^{(j-1)}$. The matrix form (\ref{matrixTjEven}) tells us that the first row of $T^{(j)}$ has only one nonzero entry, the first entry 
${\overset{\circ}{w}}^{(j-J_1)}_0$, so we know that 
$$ \left(T^{(j)} h^{(j-1)} (x)\right)_1 ={\overset{\circ}{w}}^{(j-J_1)}_0 \left(\Pi_{i=1}^{j-J_1 -1} \overset{\circ}{w}^{(i)}_0\right) \xi \cdot x  + {\overset{\circ}{w}}^{(j-J_1)}_0 B^{(j-1)}. $$ 
From the choice of the bias vector $b^{(j)}$, we find that in both cases $j-1=J_1$ and $j-1>J_1$,
$b^{(j)}_1 = {\overset{\circ}{w}}^{(j-J_1)}_0 B^{(j-1)} - B^{(j)}$. It follows that 
$$ \left(h^{(j)} (x)\right)_1 = \sigma\left({\overset{\circ}{w}}^{(j-J_1)}_0 \left(\Pi_{i=1}^{j-J_1 -1} \overset{\circ}{w}^{(i)}_0\right) \xi \cdot x + B^{(j)}\right) = \left(\Pi_{i=1}^{j-J_1} \overset{\circ}{w}^{(i)}_0\right) \xi \cdot x + B^{(j)}, $$
where we have used the fact that $\left|\left(\Pi_{i=1}^{j-J_1} \overset{\circ}{w}^{(i)}_0\right) \xi \cdot x\right| \leq B^{(j)}$. This verifies the first entry of (\ref{iterationhjrevise}). 

Now we consider the even entries of (\ref{iterationhjrevise}).
Observe from (\ref{matrixTjEven}) that the $2i$-th row of $T^{(j)}$ has all odd entries to be zero.
Then
$$ \left(T^{(j)} h^{(j-1)} (x)\right)_{2i} = \sum_{1\leq k \leq d_{j-1}/2} {\overset{\circ}{w}}^{(j-J_1)}_{i-k} \left(h^{(j-1)} (x)\right)_{2k}.  $$
For $i=1, 2, \ldots, D_{j-J_1} = d+ (j-J_1) S$, we see from $2\left(d+ (j-J_1) S\right) = 2 d + (j-J_1) s \leq d_{j-1}$ and the hypothesis assumption $\left(h^{(j-1)} (x)- \delta_{j-1, J_1} B^{(j-1)} {\bf 1}_{d_{j-1}}\right)_{2k} = \left(H^{(j-J_1 -1)}\right)_k$ that
$$ \left(T^{(j)} h^{(j-1)} (x) - b^{(j)}\right)_{2i} =\left(T^{(j)} \left(h^{(j-1)} (x) -\delta_{j-1, J_1} B^{(j-1)} {\bf 1}_{d_{j-1}}\right) - \beta^{(j)}\right)_{2i} = \sum_{k=1}^i {\overset{\circ}{w}}^{(j-J_1)}_{i-k} \left(H^{(j-J_1 -1)}\right)_k - \beta^{(j)}_{2 i}.  $$
It follows from $\beta^{(j)}_{2 i} = \overset{\circ}{b}^{(j-J_1)}_i$ that the $2i$-th entry of $h^{(j)} (x)$ equals 
$$ \left(h^{(j)} (x)\right)_{2i} = \left(\sigma \left({\overset{\circ}{T}}^{(j-J_1)} H^{(j-J_1-1)}(x)  - \overset{\circ}{b}^{(j-J_1)}\right)\right)_i$$
which is exactly $\left(H^{(j-J_1)}(x)\right)_i.$

For $i> D_{j-J_1} = d+ (j-J_1) S$, we have
$$ \left(T^{(j)} h^{(j-1)} (x)- b^{(j)}\right)_{2i} = \sum_{k=1}^{D_{j-J_1}} {\overset{\circ}{w}}^{(j-J_1)}_{i-k} \left(H^{(j-J_1 -1)}\right)_k- \beta^{(j)}_{2 i}.$$
The first term on the right-hand side is bounded by $\left\|{\overset{\circ}{w}}^{(j-J_1)}\right\|_1 \left\|H^{(j-J_1-1)}\right\|_\infty.$
Hence by the choice of $\beta^{(j)}_{2 i}$ we obtain $\left(h^{(j)} (x)\right)_{2i} = 0$. 
This verifies (\ref{iterationhjrevise}) for the even entries. 

As for the other odd entries with indices $2i+1$ and $i\geq 1$, we notice that the corresponding row of $T^{(j)}$ has all even entries to be zero.
But the only nonzero odd entry of $h^{(j-1)} - \delta_{j-1, J_1} B^{(j-1)} {\bf 1}_{d_{j-1}}$ is the first one.
Hence
\begin{eqnarray*}
\left(T^{(j)} h^{(j-1)} (x)-b^{(j)}\right)_{2i+1} &=&\left(T^{(j)} \left(h^{(j-1)} (x)- \delta_{j-1, J_1} B^{(j-1)} {\bf 1}_{d_{j-1}}\right)-\beta^{(j)}\right)_{2i+1} \\
&=&{\overset{\circ}{w}}^{(j-J_1)}_{i}  \left(\left(\Pi_{i=1}^{j-J_1-1} {\overset{\circ}{w}}^{(i)}_0\right) \xi \cdot x  + \left(1-\delta_{j-1, J_1}\right) B^{(j-1)}\right) -\beta^{(j)}_{2i+1}.
\end{eqnarray*}
Note that the first term on the right-hand side is bounded by $2 B^{(j)}$. By the choice of $\beta^{(j)}_{2i+1}$, we see that $\left(h^{(j)} (x)\right)_{2i+1} =0$.

Combining all the above three cases verifies our claim for $j$, and proves the proposition.
\end{proof}

At the end of the second group of layers, the width is $d_{J_1 +J_2} = d_{J_1} + J_2 s$ and
\begin{equation}
\label{initialexplicit}h^{(J_1 +J_2)} (x) = \left[
\left(\Pi_{i=1}^{J_2} \overset{\circ}{w}^{(i)}_0\right) \xi \cdot x  + B^{(J_1 +J_2)},
\left(H^{(J_2)} (x)\right)_1,
0,\cdots, 0, \left(H^{(J_2)} (x)\right)_{D_{J_2}}, 0 ,\cdots ,
0\right]^T.
\end{equation}

\subsection{The Third Step: Expanding DCNN to replicate linear feature}
\label{sec:construct-cnn-for-linear}

To this end, we have constructed a DCNN of depth $J_1+J_2$ where the $(J_1+J_2)$-th layer is given by $h^{(J_1 +J_2)}$ in (\ref{initialexplicit}). For convenience, we denote
 \begin{equation}\label{vectorC}
    \widehat{C}=\left[B^{(J_1 +J_2)},\ 0,\cdots ,0\right]^T \in \RR^{d_{J_1+J_2}}. 
 \end{equation}
Then the function vector \begin{equation}\label{vectorU}
  \widehat{U}(x):= h^{(J_1 +J_2)}(x) - \widehat{C}   
\end{equation} has the linear feature $\left(\Pi_{i=1}^{J_2} \overset{\circ}{w}^{(i)}_0\right) \xi \cdot x$ as the first component and the entries of $H^{(J_2)} (x)$ as some even components.
Recall that $\left(\Pi_{i=1}^{J_2} \overset{\circ}{w}^{(i)}_0\right) \xi \cdot x \leq \left(\Pi_{i=1}^{J_2} \overset{\circ}{w}^{(i)}_0\right)B^{(J_1)} = B^{(J_1 +J_2)}$. All the components of $\widehat{U}(x)$ are 
bounded uniformly by 
\begin{equation}\label{boundB0C}
 {\mathcal B}^{(0)}:=\max\left\{B^{(J_1 +J_2)}, \left\|H^{(J_2)}\right\|_\infty\right\}>0. \end{equation}
 
In this subsection, we shall apply Proposition \ref{generalCNN} to replicate the linear feature $\xi \cdot x$ for $N$ times by expanding our DCNN to have depth $J_1 + J_2 + J_3$, where $J_3$ is a positive integer to be determined later. To do so, we choose a sequence $W$ so that it induces a convolutional matrix $T^W$ (given in (\ref{matrixTW})) that consists of $N$ blocks of the identity matrix as 
\begin{equation}\label{TWspecialI}
 \left[\begin{array}{c}
I \\
I \\
\vdots \\
I 
\end{array}\right], 
 \end{equation}
 where $I$ denotes the $d_{J_{1} +J_2} \times d_{J_{1} +J_2}$ identity matrix.
With this matrix, the function vector $T^W \widehat{U}(x)$  contains the linear feature $\xi \cdot x$ in $N$ components and can be later used to generate ridge functions 
$\sigma\left(\eta^{*} \xi \cdot x  - t_i\right)$ required in our interpolating scheme 
(\ref{interpformBrief}). To obtain our desired form of $T^W$ given by (\ref{TWspecialI}), we can define the sequence $W$ as follows.

Let $N$ be an odd integer to be determined later. Take a sequence $W$ supported on $\{0, \ldots, (N-1) d_{J_{1} +J_2}\}$ given by
\begin{equation}\label{Wsequencesecond}
W_{i} = \left\{\begin{array}{ll}
1, & \hbox{if} \ i\in\{kd_{J_{1} +J_2}\}_{k=0}^{N -1}, \\
0, & \hbox{otherwise.}\end{array}\right.
\end{equation}
We define its symbol $\widetilde{W}$ as a polynomial on $\CC$ given in the wavelet literature \cite{daubechies1992ten} by 
$$ \widetilde{W} (z) = \sum_{j=0}^\infty W_j z^j = \sum_{k=0}^{N-1} z^{k d_{J_{1} +J_2}} =\frac{1-z^{Nd_{J_{1} +J_2}}}{1-z^{d_{J_{1} +J_2}}}, \qquad z\in \CC. $$
It has $(N -1) d_{J_{1}+J_2}$ complex roots
$$ e^{i 2 \pi\frac{\ell + j N}{N d_{J_{1}+J_2}}}, \qquad \ell=1, \ldots,  N-1, \ j=0, \ldots, d_{J_{1}+J_2}  -1 $$
appearing in complex conjugate pairs.
Applying the procedure for convolutional factorization described in Lemma \ref{filterdecom},
with an even filter length $s$,
we can find explicit expressions for the filters $\{w^{(j)}\}_{j=J_1 +J_2 +1}^{J_1 +J_2 +J_3}$ supported in $\{0, \ldots, s\}$, with
$J_3 = \left\lceil \frac{(N-1) d_{J_{1}+J_2}}{s}\right\rceil$ and $\|w^{(j)}\|_1 \geq 1$ such that \begin{equation}\label{filterW1}
    W  = w^{(J_1 + J_2 + J_3)}{*}\cdots {*}w^{(J_1 + J_2 +2)}{*}w^{(J_1 + J_2 +1)}.
\end{equation}

Recall that $D_{J_2} =d + J_2 S$ and $d_{J_1 + J_2} = d_{J_1} + J_2 s \geq 2 d+1 + 2 J_2 S > 2 D_{J_2}$. For $j> J_1 +J_2$, denote 
$$B^{(j)} = \left(\Pi_{p=I_1 +J_2 +1}^j \|w^{(p)}\|_1\right) {\mathcal B}^{(0)}. $$
The following proposition tells us that we can add $J_3$ well-defined convolutional layers to our previously obtained DCNN to get a DCNN of depth $J_1 + J_2 + J_3$. This DCNN shall output a vector that contains  $\xi \cdot x$ in $N$ components. 

\begin{proposition}\label{realizinginterpolation}
Let $n\in\NN$, $N \geq 3 n$ be an odd integer, $t_1 < t_2 < \cdots < t_{3n}$, and 
$W$ be the sequence given by (\ref{Wsequencesecond}) with ${\mathcal V} =(N-1) d_{J_1 +J_2}$. Take $K=d_{J_1+J_2},\ J^{*}  = J_3 :=\lceil \frac{(N-1) d_{J_1 +J_2}}{s-1}\rceil$, the initial layer $\hat{h}^{(0)} (x)=  \widehat{U}(x)+ \widehat{C}$ (with $ \widehat{U}(x)$ given by (\ref{vectorU}), $\widehat{C}$ given by (\ref{vectorC})) and the constant ${\mathcal B}^{(0)}$ given by (\ref{boundB0C}). Construct the CNN layers $\left\{h^{(j)}(x): \RR^d \to \RR^{{d_j}}\right\}_{j=J_1 +J_2+1}^{J_1 + J_2 +J_3}$ as in Proposition \ref{generalCNN} with the last bias vector
 $$ b^{(J_1 + J_2 +J_3)}= B^{(J_1 +J_2 +J_3 -1)} T^{(J_1 + J_2 +J_3)}{\bf 1}_{d_{J_1 + J_2 +J_3 -1}} + \beta $$
given in terms of the vector $\beta \in \RR^{d_{J_1 + J_2 +J_3}}$ by
$$ \left[\beta_{(i-1) d_{J_1 + J_2} +1}\right]_{i=1}^{3n} = \left[|\overset{*}{w}|  t_i\right]_{i=1}^{3n} $$
and
$$ \beta_i = \left\{\begin{array}{ll}
- B^{(J_1 +J_2 +J_3)}, & \hbox{if} \  i\in\{2k\}_{k=1}^{D_{J_2}}, \\
B^{(J_1 +J_2 +J_3)}, & \hbox{if} \  i\not\in\{2k\}_{k=1}^{D_{J_2}} \cup \left\{(i-1) d_{J_1 + J_2} +1\right\}_{i=1}^{3n},
\end{array}\right. $$
where $\overset{*}{w} := \Pi_{i=1}^{J_2} \overset{\circ}{w}^{(i)}_0$.
Then we have
$$\left(h^{(J_1 + J_2 +J_3)} (x)\right)_{i} = \left\{\begin{array}{ll}
|\overset{*}{w}| \sigma\left(\hbox{sgn}(\overset{*}{w}) \xi \cdot x  - t_k\right), &
\hbox{if} \ i= (k-1) d_{J_1 + J_2} +1 \ \hbox{with} \ k\in \{1, \ldots, 3n\}, \\
\left(H^{(J_2)} (x)\right)_k + B^{(J_1 +J_2 +J_3)}, &
\hbox{if} \ i=2k \ \hbox{with} \ k\in \{1,  \ldots, D_{J_2}\}, \\
0, & \hbox{otherwise.}
\end{array}\right. $$
There is only one free parameter $ {\mathcal B}^{(0)} = \max\left\{B^{(J_1 +J_2)}, \left\|H^{(J_2)}\right\|_\infty\right\}$ in this construction. 
\end{proposition}

\begin{proof}
Observe that $B^{(j)}\geq \left(\Pi_{p=J_1 +J_2 + 1}^{j} \|w^{(p)}\|_1\right) \left\|\widehat{h}^{(J_1 +J_2)}\right\|_\infty$ for $j=J_1+J_2+1, \ldots, J_1+J_2+J_3$.

Consider $\widehat{U}(x)+ \widehat{C}=h^{(J_{1} +J_2)} (x)$ given by (\ref{initialexplicit}) and the filters $\{w^{(j)}\}_{j=J_1 +J_2+1}^{J_1 + J_2 +J_3}$
explicitly constructed above satisfying (\ref{Wsequencesecond}) and (\ref{filterW1}) with $N$.
Define the CNN layers $\left\{h^{(j)}(x): \RR^d \to \RR^{{d_j}}\right\}_{j=J_1 +J_2+1}^{J_1 + J_2 +J_3}$
with these filters and bias vectors are chosen to be
$$b^{(J_1 +J_2 +1)}=
T^{(J_1 +J_2 +1)}\widehat{C} -
B^{(J_1 +J_2 +1)}  {\bf 1}_{d_{J_1+J_2+1}},$$
$\{b^{(J_1+J_2+j)}\}_{j=2}^{J_3-1}$ given by (\ref{biasbjD}) with $D=d_{J_{1} +J_2}, J^{*} = J_3, \mathcal{B}^{(0)}$ given by (\ref{boundB0C}) and $b^{(J_1 + J_2 +J_3)}$ defined above.

According to Proposition \ref{generalCNN}, we have 
\begin{eqnarray*}
T^{(J_1 + J_2 +J_3)} \left(h^{(J_1 + J_2 +J_3 -1)} (x)\right) = T^W \widehat{U}(x)
+ B^{(J_1 +J_2 +J_3 -1)} T^{(J_1 + J_2 +J_3)}{\bf 1}_{d_{J_1 + J_2 +J_3 -1}}.
\end{eqnarray*}
Putting the choice of  $b^{(J_1 + J_2 +J_3)}$ and the special form (\ref{TWspecialI}) of the matrix $T^W$ into the above expression, we obtain
\begin{eqnarray*}
 h^{(J_1 + J_2 +J_3)}(x) &=& T^{(J_1 + J_2 +J_3)} \left(h^{(J_1 + J_2 +J_3 -1)} (x)\right) - b^{(J_1 + J_2 +J_3)} \\
 &=& \left[\widehat{U}(x)^T \ \widehat{U}(x)^T \ \cdots \ \widehat{U}(x)^T\right]^T -\beta.
\end{eqnarray*}

Observe that each component of $\widehat{U}(x)$ is bounded by ${\mathcal B}^{(0)}\leq B^{(J_1 +J_2 +J_3)}$ and
$\beta_i = B^{(J_1 +J_2 +J_3)}$ for $i\not\in\{2k\}_{k=1}^{D_{J_2}} \cup \left\{(i-1) d_{J_1 + J_2} +1\right\}_{i=1}^{3n}$, we know that $\left(h^{(J_1 + J_2 +J_3)} (x)\right)_i$ with such an index $i$ is zero.

For  $k\in\{1, 2, \ldots, D_{J_2}\}$, $\left(h^{(J_1 + J_2 +J_3)} (x)\right)_{2k}$ is
$$  \sigma\left(\left(H^{(J_2)} (x)\right)_k + B^{(J_1 +J_2 +J_3)}\right) = \left(H^{(J_2)} (x)\right)_k + B^{(J_1 +J_2 +J_3)}. $$

For $i\in \{1, 2, \ldots, 3n\}$, we have
$$ \left(h^{(J_1 + J_2 +J_3)} (x)\right)_{(i-1) d_{J_1 + J_2} +1} =|\overset{*}{w}| \sigma\left(\hbox{sgn}(\overset{*}{w}) \xi \cdot x  - t_i\right). $$
This verifies the stated expression for $\left(h^{(J_1 + J_2 +J_3)} (x)\right)_{i}$. 
The proof is complete.
\end{proof}

\subsection{Achieving interpolations using ridge functions}
\label{sec:achieve-interpolate}

The hypothesis space induced from the DCNN of depth $J_1 +J_2 +J_3$ constructed above  is 
\begin{align*}
{\mathcal H^*} &= \hbox{span}\left\{c\cdot h^{(J_1 + J_2 +J_3)} (x) + a: c\in\RR^{d_{J_1 +J_2 +J_3}}, a\in \RR\right\} \\
& =\left\{\alpha \cdot H^{(J_2)} (x) + \gamma \cdot \left(\sigma\left(\hbox{sgn}\left(\overset{*}{w}\right) \xi \cdot x  - t_k\right)\right)_{k=1}^{3n} +a: \alpha \in \RR^{D_{J_2}}, \gamma \in \RR^{3n}, a\in \RR \right\}.    
\end{align*}
The number of free parameters of an output function from this hypothesis space equals
$$
d_{J_1+J_2+J_3}+1+
J_1(s+2)+1+1.
$$

We are now in the position to prove Theorem \ref{MainResult}. Let us first recall what Theorem 1 suggests.
Suppose that there is a DCNN $\left\{H^{(j)}(x): \RR^d \to \RR^{{D_j}}\right\}_{j=0}^{J_2}$ of depth $J_2$  with filter length $S \in\NN$
given in Section \ref{sec:deepening} satisfies the following (same as \eqref{pre-error}):
\begin{equation*}
\inf_{f\in {\mathcal H}_{J_2,S}} \left\|\pi_M f - f_\rho\right\|_\rho < E_{J_2},
\end{equation*}
where
$\{E_{J_2}>0\}_{J_2\in\NN}$ is an error sequence achieved by this CNN, which may depend on the confidence level.
Theorem \ref{MainResult} shows that this bound of excess generalization error can be maintained by some $f\in {\mathcal H^*}$ (DCNN of depth $J_1+J_2+J_3$), while satisfying the interpolation condition (\ref{interpolation}).
\bigskip

\begin{proof}[Proof of Theorem \ref{MainResult}]
Since a convolutional neural network depends on the filter sequence continuously, we know that the generalization error bound (\ref{pre-error}) can be achieved by an output function $f^{*}(x) =c^{*} \cdot H^{(J_2)} (x) + a^{*}$ with $c^{*} \in\RR^{D_{J_2}}, a^{*} \in\RR$ produced by a DCNN $\left\{H^{(j)}(x): \RR^d \to \RR^{{D_j}}\right\}_{j=0}^{J_2}$ of depth $J_2$ satisfying
$$ {\overset{\circ}{w}}^{(j)}_0 \not= 0, \qquad \forall j=1, \ldots, J_2. $$
Under this condition, we have $\overset{*}{w} = \Pi_{i=1}^{J_2} \overset{\circ}{w}^{(i)}_0 \not= 0$.

From the assumption that $\rho_\Omega$ has no positive mass at any point $x\in\Omega$, we know that the points in the set $\left\{x^i\right\}_{i=1}^n$ are distinct almost surely.

When $\left\{x^i\right\}_{i=1}^n$ are distinct, we see that for any $i\not= j \in \{1, \ldots, n\}$, $(x^i - x^j)^\perp$ is a subspace of $\RR^d$ of co-dimension $1$, where for a nonzero vector $\eta \in\RR^d$, $\eta^\perp =\{x\in \RR^d:  \eta \cdot x=0\}$
denotes the subspace of $\RR^d$ perpendicular to $\eta$.
Also, for each $i\in\{1, \ldots, n\}$, the set $x^i + \eta^\perp$ has $\rho_\Omega$ measure $0$ for almost every $\eta \in \RR^d$ with respect to the Lebesgue measure.
Therefore, there exists some nonzero vector $\xi \in\RR^d$ such that
$$\hbox{sgn}(\overset{*}{w}) \xi \cdot (x^i - x^j) \not=0, \qquad \forall i\not= j \in \{1, \ldots, n\}$$
and
$$  \rho_\Omega \left(x^i + \left(\hbox{sgn}(\overset{*}{w})\xi\right)^\perp\right) =0, \qquad \forall i \in \{1, \ldots, n\}. $$
This implies that the set $\{u_i :=\hbox{sgn}(\overset{*}{w}) \xi \cdot x^i\}_{i=1}^n$ contains distinct real numbers.

Now we can construct an interpolator $f$ satisfying the condition $f(x^i)=y^i, \ \forall i=1,\ldots, n$.
Denote
$$\epsilon^* = \min_{i\not= j \in \{1, \ldots, n\}} \left\{\frac{1}{2} \left|\xi \cdot (x^i - x^j)\right|\right\}. $$
For $\epsilon \in (0, \epsilon^*)$, recall the hat function  $\phi =\phi^{(\epsilon)}: \RR \to \RR$ given by
$$\phi(u) = \frac{1}{\epsilon} \left\{\sigma\left(u+ \epsilon\right) - \sigma\left(u\right)\right) - \left(\sigma\left(u\right) - \sigma\left(u- \epsilon\right)\right\}. \qquad u\in\RR. $$
It is supported on  $[-\epsilon, \epsilon]$ and equal to $1$ at $0$. The interpolator is given by
\begin{equation}\label{interpform}
f(x) =f^{*}(x) +  \sum_{\ell=1}^n \left(y^\ell - f^{*}(x^\ell)\right) \phi\left(\hbox{sgn}(\overset{*}{w}) \xi \cdot x  - u_\ell\right).
\end{equation}
From the definition of the hat function $\phi$, we can see that $f\in \mathcal H^*$ with
$\{t_i: i=1, \ldots, 3n\} =\left\{u_\ell -\epsilon: \ell=1, \ldots, n\right\} \cup \left\{u_\ell: \ell=1, \ldots, n\right\} \cup \left\{u_\ell+\epsilon: \ell=1, \ldots, n\right\}$.

The function $f$ is an interpolator satisfying $f(x^i)=y^i \ \forall i$ and thereby lies in $\mathcal{H}_{int,J_1+J_2+J_3,S}$
because for each $i \in\{1, \ldots, n\}$, we have
$\hbox{sgn}(\overset{*}{w})  \xi \cdot x^i = u_{i}$  implying that
$$ \phi\left(\hbox{sgn}(\overset{*}{w}) \xi \cdot x^i  - u_\ell\right) = \phi\left(u_{i}  - u_\ell\right) =\begin{cases}
                 1 & \text{if } i=\ell,\\
                 0 & \text{if } i \neq \ell,
    \end{cases}  , \qquad \ell \in\{1, \ldots, n\}. $$
Hence, we have $f(x^i) = f^{*}(x^i) +  y^i - f^{*}(x^i) =y^i$.

Finally, we estimate the excess generalization error of $\pi_M f$.
Notice that the function $\phi\left(\hbox{sgn}(\overset{*}{w}) \xi \cdot x  - u_\ell\right)$ is zero unless $\hbox{sgn}(\overset{*}{w}) \xi \cdot x  - u_\ell \in (-\epsilon, \epsilon)$. It follows that the function
$$ \sum_{\ell=1}^n \left(y^\ell - f^{*}(x^\ell)\right) \phi\left(\hbox{sgn}(\overset{*}{w}) \xi \cdot x  - u_\ell\right) $$
on the right-hand side of (\ref{interpform}) is supported on the set
$$ X_\epsilon := \cup_{\ell=1}^n \left\{x\in \RR^d: \left|\hbox{sgn}(\overset{*}{w}) \xi \cdot (x  - x^\ell)\right| <\epsilon\right\}. $$
Hence we find from the expression (\ref{interpform}) of $f$ that $\pi_M f(x) = \pi_M f^{*}(x)$ for $x\not\in X_\epsilon$ while
$\left|\pi_M f(x) - \pi_M f^{*}(x)\right| \leq 2M$ for $x\in X_\epsilon$. Thus, we have
\begin{eqnarray*}
\left\|\pi_M f- \pi_M f^{*}\right\|_\rho &=& \left\{\int_{X_\epsilon} \left|\pi_M f(x) - \pi_M f^{*}(x)\right|^2 d \rho_\Omega\right\}^{1/2}
\leq 2M \left\{\rho_\Omega \left(X_\epsilon\right)\right\}^{1/2} \\
&& \to 2M \left\{\rho_\Omega \left(\cup_{\ell=1}^n \left\{x^\ell + \left(\hbox{sgn}(\overset{*}{w})\xi\right)^\perp\right\}\right)\right\}^{1/2} =0
\end{eqnarray*}
as $\epsilon \to 0^+$.
Then, our estimate follows from the triangle inequality $\left\|\pi_M f-f_\rho\right\|_\rho \leq  \left\|\pi_M f-\pi_M f^{*}\right\|_\rho + \left\|\pi_M f^{*}-f_\rho\right\|_\rho$ and
 (\ref{pre-error}) for $f^{*}$. 
 The proof of Theorem \ref{MainResult} is complete. 
\end{proof}

Now that we completed the proof of Theorem $\ref{MainResult}$, let us move on to the proof of Theorem $\ref{MainResult2}$.

\section{Learning Rates of DCNNs} \label{Proof2}

In this section, we derive learning rates of DCNNs for proving Theorem \ref{MainResult2}. In other words, we establish an upper bound of the excess error $\varepsilon(\pi_M f_{D,J,S}) - \varepsilon (f_\rho) =\left\|\pi_M f_{D,J,S} - f_\rho\right\|_2^2$. Our analysis is based on an error decomposition (Subsection \ref{errordecomposition}), followed by bounding the approximation error and the sampling error, respectively (Subsection \ref{sampleerror}).
Afterward, we bound the number of layers and achieved the desired learning rates (Subsection \ref{optimallayer}). Our method is different from the approaches in the existing literature of generalization analysis of DCNNs \cite{mao2021theory, fang2022optimal}. By using a pseudo-dimension estimate of the induced hypothesis space, we no longer require the free parameters to be bounded. 
Lastly, by applying the results in Theorem \ref{MainResult2}, we are able to derive the proof of Theorem \ref{MainResult3} (Subsection \ref{proof3}). 

Note that the filter length specified in Theorem \ref{MainResult2} is $2 \leq S \leq d$.

\subsection{Error decomposition and estimating the approximation error} \label{errordecomposition}

We aim to find the upper bounds of the excess error: 
$$
\varepsilon(\pi_M f_{D,J,S}) - \varepsilon (f_\rho) =\left\|\pi_M f_{D,J,S} - f_\rho\right\|_2^2 =  \int_{Z} (\pi_M f_{D,J,S}(x)-f_\rho(x))^2 \,d\rho,
$$
where $\pi_M f_{D,J,S}$ is the truncated global minimizer of (\ref{ERM}) and $f_\rho$ is the regression function.
Let $f_{\mathcal{H}_{J,S}}$ be any function in the hypothesis space $\mathcal{H}_{J,S}$ defined by (\ref{hypothesis}).
To find such an upper bound, we adopt the following error decomposition.

\begin{lemma} \label{bound}
Let  $f_{D,J,S}$ be defined by (\ref{ERM}) and $f_{\mathcal{H}_{J,S}}$ be any function in the hypothesis space $\mathcal{H}_{J,S}$ defined by (\ref{hypothesis}).
Then there holds
\begin{align}
&\varepsilon(\pi_M f_{D,J,S}) - \varepsilon (f_\rho) 
\leq
 \left\{(\varepsilon(\pi_M f_{D,J,S}) - \varepsilon (f_\rho))-(\varepsilon_D(\pi_M f_{D,J,S}) - \varepsilon_D (f_\rho))\right\}   \label{item4} \\
 & \qquad +  \left\{(\varepsilon_D(f_{\mathcal{H}_{J,S}})-\varepsilon_D (f_\rho))-(\varepsilon(f_{\mathcal{H}_{J,S}})-\varepsilon(f_\rho))\right\}
 + \left\{\varepsilon(f_{\mathcal{H}_{J,S}})-\varepsilon(f_\rho)\right\} \nonumber
 =: \mathcal{A}_1 + \mathcal{A}_2+ \mathcal{A}_3.
\end{align}
\end{lemma}
\begin{proof}
We express the excess error by inserting empirical error terms as follows
\begin{align*}
\varepsilon(\pi_M f_{D,J,S}) - \varepsilon (f_\rho) &=
\{\varepsilon(\pi_M f_{D,J,S}) - \varepsilon (f_\rho)\}-\{\varepsilon_D(\pi_M f_{D,J,S}) - \varepsilon_D (f_\rho)\}   \\
 & +\{\varepsilon_D(\pi_M f_{D,J,S}) - \varepsilon_D (f_\rho)\} - \{\varepsilon_D(f_{\mathcal{H}_{J,S}})-\varepsilon_D (f_\rho)\}    \\
 &+ \{\varepsilon_D(f_{\mathcal{H}_{J,S}})-\varepsilon_D (f_\rho)\}-\{\varepsilon(f_{\mathcal{H}_{J,S}})-\varepsilon(f_\rho)\}
+ \{\varepsilon(f_{\mathcal{H}_{J,S}})-\varepsilon(f_\rho)\}.
\end{align*}
As both $y^i$ and $\pi_M f_{D,J,S}(x^i)$ are values on the interval $[-M, M]$ for each $i$, we know that $\left(\pi_M f_{D,J,S}(x^i) - y^i\right)^2 \leq
\left(f_{D,J,S}(x^i) - y^i\right)^2$ which yields  $\varepsilon_D(\pi_M f_{D,J,S}) \leq \varepsilon_D(f_{D,J,S})$.
Note that both $f_{D,J,S}$ and $f_{\mathcal{H}_{J,S}}$ lie in the space $\mathcal{H}_{J,S}$.
By definition,
$f_{D,J,S}$ minimizes the empirical risk $\varepsilon_D (f)$ over $\mathcal{H}_{J,S}$.
It follows that
\begin{align*}
\varepsilon_D(\pi_M f_{D,J,S}) - \varepsilon_D (f_\rho)- \{\varepsilon_D(f_{\mathcal{H}_{J,S}})-\varepsilon_D (f_\rho)\}
&=\varepsilon_D(\pi_M f_{D,J,S}) - \varepsilon_D(f_{\mathcal{H}_{J,S}})\leq 0.
\end{align*}
The expression (\ref{item4}) is verified.
\end{proof}

To achieve the learning rates stated in Theorem \ref{MainResult2}, we are going to bound the three terms,  $\mathcal{A}_1, \mathcal{A}_2, \mathcal{A}_3$, on the right-hand side of $(\ref{item4})$ in the following.

The last term of $(\ref{item4})$, $ \mathcal{A}_3 =\varepsilon(f_{\mathcal{H}_{J,S}})-\varepsilon(f_\rho) = \left\|f_{\mathcal{H}_{J,S}}-f_\rho\right\|_2^2 \leq \left\|f_{\mathcal{H}_{J,S}}-f_\rho\right\|_{C(\Omega)}^2$, is known as the approximation error, where the last inequality holds due to the fact that the marginal distribution $\rho_\Omega$ has measure $1$. 
It has been estimated in \cite[Theorem 2]{zhou2020universality} as follows. 

\begin{lemma}\label{Theorem B}
Let $2 \leq S\leq d$ and $\Omega \subseteq [-1, 1]^d$. If $J \geq 2d/(S-1)$ and $f_\rho=F|_{\Omega}$ with $F \in W_2^{r} ({\mathbb R}^d)$ and an integer index $r>2 + d/2$,
then there exist ${\bf w}$, ${\bf b}$ with (\ref{baisform}) valid for layers $1, 2, \ldots J-1$ and $f_{\mathcal{H}_{J,S}} \in \mathcal{H}_{J,S}$ such that
\begin{equation}\label{approxerrorest}
\mathcal{A}_3 \leq \|f_{\mathcal{H}_{J,S}} -f_\rho\|_{C(\Omega)} \leq c \left\|F\right\|_{W_2^{r}} \sqrt{\log J} J^{-\frac{1}{2} - \frac{1}{d}},
\end{equation}
where $c$ is an absolute constant and $\left\|F\right\|_{W_2^{r}}$ denotes the Sobolev space norm of $F$.
\end{lemma}

The second term of (\ref{item4}), $\mathcal{A}_2=\varepsilon_D(f_{\mathcal{H}_{J,S}})-\varepsilon_D (f_\rho)-\{\varepsilon(f_{\mathcal{H}_{J,S}})-\varepsilon(f_\rho)\}$, can be estimated by the Bernstein inequality \cite{bernstein1927extension} (see, e.g., \cite[Lemma A.2]{bosq2000linear}).

\begin{lemma} \label{bounditem2}
For any $0< \delta <1$, with probability at least $1-\frac{\delta}{2}$, we have
\begin{align*}
\mathcal{A}_2&=\{\varepsilon_D(f_{\mathcal{H}_{J,S}})-\varepsilon_D (f_\rho)\}-\{\varepsilon(f_{\mathcal{H}_{J,S}})-\varepsilon(f_\rho)\} \\
&\leq 4\log\left(\frac{2}{\delta}\right) \left\|f_{\mathcal{H}_{J,S}}-f_\rho\right\|_\infty \left(\left\|f_{\mathcal{H}_{J,S}}-f_\rho\right\|_\infty  + 4M\right)/\sqrt{n}.
\end{align*}
\end{lemma}

\begin{proof}
Define a random variable $\zeta$ on $(Z, \rho)$ by $\zeta(z)=\zeta(x,y)=(f_{\mathcal{H}_{J,S}}(x)-y)^2-(f_\rho(x)-y)^2$. We have
$
\varepsilon(f_{\mathcal{H}_{J,S}})-\varepsilon(f_\rho) =\mathrm{E}[\zeta]$ and
$
\varepsilon_D(f_{\mathcal{H}_{J,S}})-\varepsilon_D (f_\rho) =\frac{1}{n}\sum_{i=1}^n \zeta(z^i).$ Hence, the second term of (\ref{item4}) can be expressed in terms of $\zeta$ as
\begin{equation}
\{\varepsilon_D(f_{\mathcal{H}_{J,S}})-\varepsilon_D (f_\rho)\}-\{\varepsilon(f_{\mathcal{H}_{J,S}})-\varepsilon(f_\rho)\} = \frac{1}{n}\sum_{i=1}^n \zeta(z^i) - \mathrm{E}[\zeta].
\end{equation}
Note that $|f_\rho(x)-y| \leq 2M$ because $|y|\leq M$ and $|f_\rho(x)|=|\mathrm{E}[y|x]| \leq M$.
We have almost surely
\begin{align*}
|\zeta(z)| &=\left|\left(f_{\mathcal{H}_{J,S}}(x)-f_\rho(x)\right) \left(f_{\mathcal{H}_{J,S}}(x) -y + f_\rho(x)-y\right)\right|\\
&\leq \left\|f_{\mathcal{H}_{J,S}}-f_\rho\right\|_\infty \left(\left\|f_{\mathcal{H}_{J,S}}-f_\rho\right\|_\infty  + 4M\right) =:B.
\end{align*}
The variance $\sigma^2$ of $\zeta$ can be bounded as  $\sigma^2 \leq \mathrm{E}[\zeta^2] \leq B^2$. Thus, by the one-sided Bernstein inequality,
for any $\eta>0$, there holds
$$\{\varepsilon_D(f_{\mathcal{H}_{J,S}})-\varepsilon_D (f_\rho)\}-\{\varepsilon(f_{\mathcal{H}_{J,S}})-\varepsilon(f_\rho)\} \leq \eta $$
with probability at least $1-\exp\left(-\frac{n\eta^2}{2(\sigma^2+\eta B/3)}\right)$. Setting this confidence bound to be $1-\frac{\delta}{2}$,
we get a quadratic equation $\frac{n\eta^2}{2(\sigma^2+\eta B/3)} = \log \frac{2}{\delta}$ for $\eta$.
The positive solution $\eta^*$ to this equation is given by
\begin{align*}
 \eta^* &= \frac{\frac{2B}{3}\log\left(\frac{2}{\delta}\right)+\sqrt{\left(\frac{2B}{3}\log\left(\frac{2}{\delta}\right)\right)^2+8n\sigma^2\log\left(\frac{2}{\delta}\right)}}{2n}\\
&\leq \frac{2B\log\left(\frac{2}{\delta}\right)}{3n} +  \frac{\sqrt{2\sigma^2\log\left(\frac{2}{\delta}\right)}}{\sqrt{n}}\\
&\leq \frac{4B\log\left(\frac{2}{\delta}\right)}{\sqrt{n}}.
\end{align*}
This verifies the desired bound and completes the proof.
\end{proof}

\subsection{Estimating the sample error}\label{sampleerror}

The first term of  (\ref{item4}), $\mathcal{A}_1 = (\varepsilon(\pi_M f_{D,J,S}) - \varepsilon (f_\rho))-(\varepsilon_D(\pi_M f_{D,J,S}) - \varepsilon_D (f_\rho))$, is the sample error.
In this subsection, we derive an upper bound for this sampling error using a concentration inequality in terms of the empirical covering numbers.
Define the empirical $L_1$ norm with respect to the sample $D$ by
\begin{equation}
\|f\|_{L_1 (D)}= \frac{1}{n}\sum_{i=1}^n |f(x^i)|.
\end{equation}
For $\epsilon >0$, denote by $\mathcal{N}_1(\epsilon, \mathcal{H}, D)$ the $\epsilon$-empirical covering number of a set of functions $\mathcal{H}$ with respect to $\|\cdot\|_{L_1 (D)}$.
More specifically, $\mathcal{N}_1(\epsilon, \mathcal{H}, D)$ is the minimal $N\in \NN$ such that there exist functions $\{f_1, \ldots, f_N\} \in \mathcal{H}$ satisfying
\begin{equation}
\min_{1\leq j \leq N} \|f-f_j\|_{L_1 (D)} \leq \epsilon, \qquad \forall f\in \mathcal{H}.
\end{equation}
The following concentration inequality with arbitrary parameters $\alpha, \beta >0$ and $0< \varepsilon\leq 1/2$ can be found in \cite{Gyorfi2002} as Theorem 11.4.

\begin{lemma}\label{concentration}
Assume $|y| \leq M$ almost surely and $M\geq 1$. Let $\alpha, \beta >0$ and $0< \varepsilon\leq 1/2$.
If $\mathcal{H}$ is a set of functions $f:\RR^d \rightarrow [-M, M]$, then with probability at least
\begin{equation}
1-14\sup_{D} \mathcal{N}_1\left(\frac{\beta \varepsilon}{20M}, \mathcal{H}, D\right)\exp\left(-\frac{\varepsilon^2 (1-\varepsilon)\alpha n}{214(1+\varepsilon)M^4}\right),
\end{equation}
there holds
\begin{align*}
\left\{\varepsilon(f) - \varepsilon (f_\rho)\right\}-\left\{\varepsilon_D(f) - \varepsilon_D (f_\rho)\right\}
\leq \varepsilon\left(\alpha + \beta + \varepsilon(f) - \varepsilon (f_\rho)\right), \qquad \forall f\in \mathcal{H}.
\end{align*}
\end{lemma}

We shall apply Lemma \ref{concentration} to the function set $\mathcal{H}= \left\{\pi_M f: f\in \mathcal{H}_{J,S}\right\}$.
To this end, we need to derive a bound for its empirical covering number by means of its empirical packing number and pseudo-dimension.

Denote by $\mathcal{M}_1(\epsilon, \mathcal{H}, D)$ the $\epsilon$-empirical packing number of a set of functions $\mathcal{H}$ with respect to $\|\cdot\|_{L_1 (D)}$.
More specifically, $\mathcal{M}_1(\epsilon, \mathcal{H}, D)$ is the maximal $N\in \NN$ such that there exists functions $\{f_1,\ldots, f_N\}\in \mathcal{H}$ satisfying
$$\|f_j-f_k\|_{L_1 (D)} \geq \epsilon, \qquad  1\leq j<k \leq N. $$
A well-known relationship between $\epsilon$-covering numbers and $\epsilon$-packing numbers asserts that
\begin{equation}\label{coverpacking}
\mathcal{M}_1 (2\epsilon, \mathcal{H}, D) \leq \mathcal{N}_1(\epsilon, \mathcal{H}, D) \leq \mathcal{M}_1(\epsilon, \mathcal{H}, D), \qquad \forall \epsilon >0.
\end{equation}

For a class of real-valued functions, such as those generated by neural networks, a natural measure of complexity that implies similar uniform convergence properties is the pseudo-dimension, introduced by  \cite{pollard1990empirical} (see, e.g., \cite[Theorem 19.2]{anthony1999neural}).
 The pseudo-dimension $Pdim(\mathcal{H})$ of $\mathcal{H}$ is the largest integer $\ell$ for which there exist $\{\xi_1,\dots,\xi_\ell\} \subset \Omega$ and $\{\eta_1,\dots,\eta_\ell\} \subset \mathbb R^\ell$ such that for any $(a_1,\dots,a_\ell)\in\{0,1\}^{\ell}$ there is some $v\in\mathcal{H}$ satisfying
$$
   \quad v(\xi_i)>\eta_i\Leftrightarrow a_i=1, \qquad \forall\ i.
$$
For a set $\mathcal{H}$ of functions from $\Omega$ to $[-M, M]$, the empirical packing numbers can be bounded in terms of the pseudo-dimension by the following inequality
found in \cite[Theorem 1]{Mendelson2003}:
\begin{equation}\label{Pdimpacking}
    \mathcal M_1(\epsilon, \mathcal{H}, D)\leq 2\left(\frac{2eM}{\epsilon}\log\left(\frac{2eM}{\epsilon} \right)\right)^{Pdim(\mathcal{H})},
\end{equation}
for any $0<\epsilon \leq M$.
Observe that $Pdim(\left\{\pi_M f: f\in \mathcal{H}_{J,S}\right\})\leq Pdim(\mathcal{H}_{J,S})$. \\

The last tool we use is an important bound for the pseudo-dimension of a deep ReLU neural network, which can be found in \cite[Theorem 7]{Bartlett2019}.
This result is valid only for piecewise polynomial activation functions
but is crucial in our study with the ReLU activation function, which is piecewise linear.
It asserts that for a ReLU neural network architecture with 
$U$ computation units (neurons) arranged in $L$ layers,
the pseudo-dimension of the hypothesis space is bounded by $c \sum_{i=1}^L W_i \log U$, where $c$ is an absolute constant and $W_i$ is the total number of parameters (weights and biases)
at the inputs to units in {\it all the layers  up to layer $i$}.
We find by the same proof that this bound still holds when we replace the parameters in weights and biases by {\it free parameters}.
This bound may be extended to some other neural networks with special structures. We state it as a theorem below (Theorem \ref{generalpseudodim}), of independent interest, and give detailed proof in Appendix A.2. 

Motivated by convolutional neural networks, which have many weight entries to be $0$ and many repeated weight and bias entries, we allow some of the parameters to be constants. We also allow some parameters within the same layer to take the same free parameter value. 
Moreover, unlike the result in \cite[Theorem 7]{Bartlett2019}, which allows neurons to have connections from any earlier layers \cite{gribonval2022approximation}, we assume that each neuron has connections only from neurons in the previous layer. 
Recall that for a DCNN, a weight matrix is a  convolutional one in which each descending diagonal from left to right shares the same value (refer to (\ref{convolution}) and (\ref{matrixTW})). Also, a bias vector has identical entries in the middle (refer to (\ref{baisform})).

Denote by $d_j$ the width, $K_j$ the number of free parameters, $A^{(j)}\in \RR^{K_j}$ the vector of free parameters in the $j$-th layer of the neural network. 
For the special case of CNN, $d_j = d + js$ and $K_j = 3s$, which consists of $s+1$ free parameters in $T^{(j)}$ and $2s-1$ in $b^{(j)}$.
Also, denote by $K$ the total number of free parameters in the neural network. 
The following theorem considers neural networks equipped with piecewise polynomial activation functions with $p + 1$ pieces.

\begin{theorem}\label{generalpseudodim}
Let $\theta, p \in \NN$ and $t_1<t_2<\cdots < t_p$ be a sequence of break points such that $\sigma: \RR \to \RR$ is continuous and $\sigma|_{(t_{i-1}, t_i)}$ is a polynomial of degree at most $\theta$ for each $ i=1,\ldots, p, p+1$ (where we set $t_0=-\infty$ and $t_{p+1} =\infty$). 
Let $J\in\NN$, $d_j, K_j \in\NN$ for $j\in\{1, \ldots, J\}$, $A^{(j)}\in \RR^{K_j}$, and $c\in \RR^{d_J}$. 
Denote $a = [A^{(1)}, \ldots, A^{(J)}, c] \in \RR^K$ with $K:=K_1 + \ldots + K_J + d_J$.
Consider a neural network $\{h^{(j)}(\cdot, a): \RR^d \to \RR^{d_j}\}_{j=1}^J$ of depth $J$ and widths $\{d_j\}_{j=1}^J$ defined with $a\in \RR^K$ by
$h^{(0)} (x, a)=x$ and iteratively
\begin{equation}\label{DNNnew}
h^{(j)} (x, a) = \sigma \left(T^{(j)} h^{(j-1)}(x, a)  - b^{(j)}\right), \qquad j=1, \ldots, J,
\end{equation}
where $T^{(j)} \in \RR^{d_{j} \times d_{j-1}}, b^{(j)}\in\RR^{d_j}$. Here, the entries of $T^{(j)}$ and $b^{(j)}$ can be divided into a collection of $K_j^\prime$ disjoint subsets with $K_j^\prime \geq K_j$. For subset $\ell$, $\ell = 1,\ldots, K_j$, all entries in this subset equal to the $\ell$-th component $A^{(j)}_\ell$ of $A^{(j)}$. For $\ell > K_j$, all entries in subset $\ell$ are equal to a fixed constant in $\RR$.
The hypothesis space generated by this network is 
\begin{equation}
{\mathcal H}_{J, \sigma} =  \hbox{span}\left\{f(x, a) =c\cdot  h^{(J)} (x, a): a \in \RR^{K}\right\}.
\end{equation}
Then we have 
\begin{equation}\label{bound-for-pseudoGeneral}
\text{Pdim}(\mathcal{H}_{J, \sigma}) \leq  J+1 + \left(d_J + \sum_{j=1}^J (J- j +2) K_j \right) \left(\log_2 (4 e R) +  \log_2 (\log_2 (2 e R))\right), 
\end{equation}
where  
$$ R:= 1 + \theta+\theta^2 + \cdots+\theta^{J} 
+ \sum_{i=1}^J d_i p (1 + \theta+\theta^2 + \cdots+\theta^{i-1}). $$
\end{theorem}

Now we apply this result to our DCNN (equipped with ReLU) given in Definition \ref{defDCNN} with the bias vectors satisfying (\ref{baisform}) for layers $1, 2, \ldots J-1$.
The proof of Lemma \ref{Lemma:Pseudo-for-DCNN} is given in Appendix A.3. 

\begin{lemma}\label{Lemma:Pseudo-for-DCNN}
For the DCNN of depth $J \geq 2$ in Definition \ref{defDCNN} with ReLU activation functions and (\ref{baisform}) valid for layers $1, 2, \ldots J-1$, we have
\begin{equation}\label{bound-for-pseudo}
Pdim(\mathcal{H}_{J,S}) \leq  c_0(J^2S+d)\log(Jd+J^2S),
\end{equation}
where $c_0$ is an absolute constant. 
\end{lemma}

We aim to find an upper bound on the first term of (\ref{item4}) as tightly as possible.
In other words, we search for the best $\alpha, \beta$, and $\varepsilon$ to minimize the upper bound in the concentration inequality presented in Lemma \ref{concentration}.

The next lemma presents an 
upper bound.

\begin{lemma} \label{bounditem1}
Assume $|y| \leq M$ almost surely and $M\geq 1$. For the DCNN of depth $J \geq 2$ in Definition \ref{defDCNN} with (\ref{baisform}) valid for layers $1, 2, \ldots J-1$, and any $0 <\delta <1$, with probability at least $1-\frac{\delta}{2}$, we have
\begin{align*}
\mathcal{A}_1 &=\{\varepsilon(\pi_M f_{D,J,S}) - \varepsilon (f_\rho)\}-\{\varepsilon_D(\pi_M f_{D,J,S}) - \varepsilon_D (f_\rho)\} \\
&\leq \frac{1}{2}\Bigl\{\frac{1}{n} +\left\{\varepsilon(\pi_M f_{D,J,S}) - \varepsilon(f_\rho)\right\}+\frac{c'_0 M^4}{n} \left\{\log(M^2 n)\left(J^2 S +d\right) \log \left(J d + J^2 S\right) + \log\left(2/\delta\right)\right\}\Bigr\},
\end{align*}
where $c'_0$ is an absolute constant. 
\end{lemma}

\begin{proof}
Take the function set $\mathcal{H}= \left\{\pi_M f: f\in \mathcal{H}_{J,S}\right\}$.
Combining (\ref{coverpacking}), (\ref{Pdimpacking}) with Lemma \ref{Lemma:Pseudo-for-DCNN},
for any $\epsilon >0$, we have
\begin{align*}
    \mathcal N_1 ( \epsilon, \mathcal{H}, D)
    &\leq 2\left(\frac{2eM}{\epsilon}\log\frac{2eM}{\epsilon} \right)^{c_0(J^2S+d)\log(Jd+J^2S)}\\
    &\leq 2\left(\frac{2eM}{\epsilon}\right)^{2c_0(J^2S+d)\log(Jd+J^2S)}.
\end{align*}

Take $\varepsilon= \frac{1}{2}$ and $\beta = \frac{1}{n}$ in Lemma \ref{concentration}. Applying the above bound, we see that the confidence bound in Lemma \ref{concentration} is
\begin{align*}
&1-14\sup_{D} \mathcal{N}_1\left(\frac{1}{40Mn}, \mathcal{H}, D\right)\exp\left(-\frac{\alpha n}{2568M^4}\right) \\
&\geq 1-28\exp\left(2c_0\log(80 e M^2 n)\left(J^2 S +d\right) \log \left(J d + J^2 S\right)-\frac{\alpha n}{2568M^4}\right).
\end{align*}
We choose $\alpha>0$ such that the above confidence bound is at least $1-\frac{\delta}{2}$, which means
\begin{equation} \label{alpha}
\alpha \geq \frac{2568 M^4}{n}\left(2c_0\log(80 e M^2 n)\left(J^2 S +d\right) \log \left(J d + J^2 S\right)+\log\left(\frac{56}{\delta}\right)\right).    
\end{equation}
This is satisfied by taking $c'_0=2^{15}(c_0+1) $ and 
\begin{equation}
\alpha^{*} = \frac{c'_0 M^4  }{n}\left(\log(M^2 n)\left(J^2 S +d\right) \log \left(J d + J^2 S\right) +\log\left(\frac{2}{\delta}\right)\right),
\end{equation}
because $\alpha^{*}$ is greater than the right hand side of (\ref{alpha}).

Substituting $\alpha, \beta$ and $\epsilon$ by the values we took into the concentration inequality in Lemma \ref{concentration}, we know that with probability at least $1-\delta/2$, there holds
$$
\left\{\varepsilon(f) - \varepsilon (f_\rho)\right\}-\left\{\varepsilon_D(f) - \varepsilon_D (f_\rho)\right\}
\leq \frac{1}{2}\left(\alpha^{*} + \frac{1}{n} + \varepsilon(f) - \varepsilon (f_\rho)\right), \qquad \forall f\in \mathcal{H},
$$
which implies
$$\{\varepsilon(\pi_M f_{D,J,S}) - \varepsilon (f_\rho)\}-\{\varepsilon_D(\pi_M f_{D,J,S}) - \varepsilon_D (f_\rho)\}
\leq \frac{1}{2}\left(\alpha^{*} + \frac{1}{n} + \varepsilon(\pi_M f_{D,J,S}) - \varepsilon (f_\rho)\right).
$$
We verify the bound and complete the proof.
\end{proof}

\subsection{Proof of Theorem \ref{MainResult2}: Achieving learning rates by bounding the number of layers} \label{optimallayer}

We are now in a position to prove Theorem \ref{MainResult2}.

\begin{proof}[Proof of Theorem \ref{MainResult2}]
Recall the number of free parameters given by (\ref{parameternum}). Since $S\leq d$, we know that this number can be bounded by
$$
3S(J-1)+S+2+2(d+JS) = S(3J-3+1+2J)+2+2d = S(5J-2)+2d+2 \leq d(5J-2)+2d+2 = 5dJ+2.
$$
Combining Lemmas \ref{bound}, \ref{Theorem B}, \ref{bounditem2} and \ref{bounditem1},  when $J \geq 2d/(S-1)$, we obtain the following bound for the excess error in terms of the number of layers $J$ with probability at least $1-\delta$:
\begin{align*}
\varepsilon(\pi_M f_{D,J,S}) - \varepsilon (f_\rho) &\leq \frac{1}{2}\left\{\varepsilon(\pi_M f_{D,J,S}) - \varepsilon (f_\rho)\right\}  \\
& + \frac{1}{2n} +\frac{c'_0 M^4}{2n} \left(\log(M^2 n)\left(J^2 S +d\right) \log \left(J d + J^2 S\right) + \log\left(2/\delta\right)\right)\\
& + \frac{4}{\sqrt{n}} \log\left(\frac{2}{\delta}\right)  (4M+1) c^2 \left\|F\right\|_{W_2^{r}}^2 \log J J^{-1 - \frac{2}{d}} \\
& + c^2 \left\|F\right\|_{W_2^{r}}^2 \log J J^{-1 - \frac{2}{d}},
\end{align*}
which implies
\begin{align*}
\varepsilon(\pi_M f_{D,J,S}) - \varepsilon (f_\rho) &\leq
\frac{1}{n} +\frac{4c'_0M^4 \log (M+1)  d (\log d)}{n} ((\log n) J^2 (\log J) + \log(2/\delta)) \\
& \quad + \left(\frac{8}{\sqrt{n}} \log(2/\delta)  (4M+1) + 2\right) c^2 \left\|F\right\|_{W_2^{r}}^2 (\log J) J^{-1 - \frac{2}{d}}\\
&\leq \frac{1}{n} + 4c'_0 M^4 \log (M+1) d (\log d)\left(\frac{(\log n) J^2 (\log J)}{n} + \frac{1}{\sqrt{n}}\right)\left(1+\frac{\log(2/\delta)}{\sqrt{n}}\right)\\
& \quad+ \left(\frac{\log(2/\delta)}{\sqrt{n}}\right)(32M+10) c^2 \left\|F\right\|_{W_2^{r}}^2\left(\frac{\log J}{J}\right)\\
&\leq \left(1+ 4c'_0 M^4 \log (M+1) d (\log d)+(32M+10) c^2 \left\|F\right\|_{W_2^{r}}^2\right)\left(1+\frac{\log(2/\delta)}{\sqrt{n}}\right)\\
& \quad \left\{\frac{(\log n) J^2 (\log J)}{n} + \frac{1}{\sqrt{n}}+\frac{\log J}{J}\right\}.
\end{align*}

This verifies the stated learning rates  in terms of $J$ at (\ref{boundJ}) with the constant $C_{M,r,f_\rho}$ given by
$$
C_{M,r,f_\rho}=1+ 4c'_0 M^4 \log (M+1) d (\log d)+(32M+10) c^2 \left\|F\right\|_{W_2^{r}}^2.
$$
We choose $J=\lceil n^\alpha\rceil$. When $n\geq (2d)^{1/\alpha}$, we see that $J \geq 2d/(S-1)$ is satisfied and thereby with probability at least $1-\delta$,
$$
\varepsilon(\pi_M f_{D,J,S}) - \varepsilon (f_\rho) \leq 6C_{M, r, f_\rho}\left(1 + \frac{\log\left(2/\delta\right)}{\sqrt{n}}\right)    \frac{(\log n)^2}{n^{\min\{1-2\alpha, \alpha\}}}.
$$
This verifies the stated learning rates in (\ref{boundwithoutJ}) with the constant $6C_{M, r, f_\rho}$.

When $n < (2d)^{1/\alpha}$, we have $\frac{(\log n)^2}{n^\alpha} \geq \frac{1}{2d}$. Then we can use a simple bound $\left|\pi_M f_{D,J,S}(x) -y\right| \leq 2 M$ and its consequence
$$\varepsilon(\pi_M f_{D,J,S}) - \varepsilon (f_\rho) \leq \varepsilon(\pi_M f_{D,J,S})
\leq 4M^2, $$
and find that with probability at least $1-\delta$,
$$ \varepsilon(\pi_M f_{D,J,S}) - \varepsilon (f_\rho) \leq 8M^2 d (\log d) \frac{(\log n)^2}{n^{\min\{1-2\alpha, \alpha\}}}.$$
This also verifies the stated learning rates at (\ref{boundwithoutJ}) with the constant $8M^2$.
The proof of Theorem \ref{MainResult2} is complete.
\end{proof}

\subsection{Proof of Theorem \ref{MainResult3}: Learning rate of interpolating DCNN} \label{proof3}
Combining Theorem \ref{MainResult} and Theorem \ref{MainResult2}, we can derive the learning rates of overparameterized DCNNs that interpolate any input data.

\begin{proof}[Proof of Theorem \ref{MainResult3}]
Take $S=s/2$, and the DCNN defined in Theorem \ref{MainResult2} as a teacher DCNN here. It has $J_2=\lceil n^\alpha\rceil$ layers for any $0<\alpha<1$, with filter length $S \in\{2, 3, \ldots, d\}$. Theorem \ref{MainResult2} states that, with probability $1-\delta$ for any $0<\delta <1$, a truncated output function $\pi_M f_{D,J,S}$ can approximate the regression function $f_\rho$ with accuracy
\begin{equation}\nonumber
\left\|\pi_M f_{D,J,S}- f_\rho\right\|_2^2 \leq \max\left\{8M^2,6C_{M, r, f_\rho}\right\} d (\log d) \left(1 + \frac{\log\left(2/\delta\right)}{\sqrt{n}}\right) \frac{(\log n)^2}{n^{\min\{1-2\alpha, \alpha\}}}.
\end{equation}

We choose $N=4n+1$ as an odd number satisfying $N \geq 3n$.
Given the learning ability of $\pi_M f_{D,J,S}$, Theorem \ref{MainResult} suggests that there exists a downsampled DCNN (student net) of depth $J_1+J_2+J_3$ with filter length $2S =s$, downsampled at the $J_1$-th layer, where $J_1=\lceil \frac{2d^2}{s-1}\rceil$, $J_3=\left\lceil \frac{(N-1) d_{J_{1}+J_2}}{s}\right\rceil=\left\lceil \frac{4n d_{J_{1}+J_2}}{s}\right\rceil=\left\lceil \frac{4n \left(d_{J_{1}}+s\lceil n^{\alpha}\rceil\right)}{s}\right\rceil$, such that with probability $1-\delta$,
\begin{equation}
\inf_{f\in {\mathcal H}_{int}} \left\|\pi_M f - f_\rho\right\|_2^2 <\max\left\{8M^2+1,6C_{M, r, f_\rho}+1\right\} d (\log d) \left(1 + \frac{\log\left(2/\delta\right)}{\sqrt{n}}\right) \frac{(\log n)^2}{n^{\min\{1-2\alpha, \alpha\}}}.
\end{equation}
The proof of Theorem \ref{MainResult3} is completed.
\end{proof}

\section{Numerical Experiment} \label{experiment}
In this section, we present our results of numerical experiments on simulated data to corroborate our theoretical findings in Theorem \ref{MainResult2}.
Our purpose is to demonstrate our theoretical findings. We do not intend to perform numerical experiments with real data.

We conduct numerical experiments for four different input data dimensions $d\in \{10, 30, 50, 100\}$. For each $d$, we simulate $n$ training data sets $\{(x^i, y^i)\}_{i=1}^n$ with $n$ varying in \\$\{100,300,500, 700, 1000,1500, 2000, 3000, 4000,
5000, 6000\}$, $x^i \in \RR^d$ being a random vector with entries uniformly distributed in $[-10,10]$, and $y^i \in \RR$ generated from the following regression model:
\begin{equation}
    y^i = \sin (\|x^i\|_2^4) + \cos(\|x^i\|_2^4) + \epsilon^i,
\end{equation}
where $\epsilon^i$ is random noise following $\mathcal{N}(0, 0.01)$. Corresponding to each training data set, we simulate a test data set $\{(x^i_{\text{test}}, y^i_{\text{test}})\}_{i=1}^{2000}$ where $x^i_{\text{test}} \in \RR^d$ has entries uniformly distributed in $[-10,10]$ and $ y^i_{\text{test}}=\sin (\|x^i_{\text{test}}\|_2^4) +\cos(\|x^i_{\text{test}}\|_2^4).$ This choice of the regression function is bounded in $[-2,2]$.

To run the experiments, we adopt the DCNNs following the structure defined in Definition \ref{defDCNN}. We fix the filter length and depth of the network to be $S=2$ and $J=\lceil n^\frac{1}{3}\rceil$ respectively. For convenience, we neglect the special structure of the bias vector stated in equation (\ref{baisform}). We train our network using the famous Adam optimization algorithm with constant step sizes, and free parameters (weights and bias) initialized by the default TensorFlow values.

\begin{figure}[h] 
    \includegraphics[width=19cm, center]{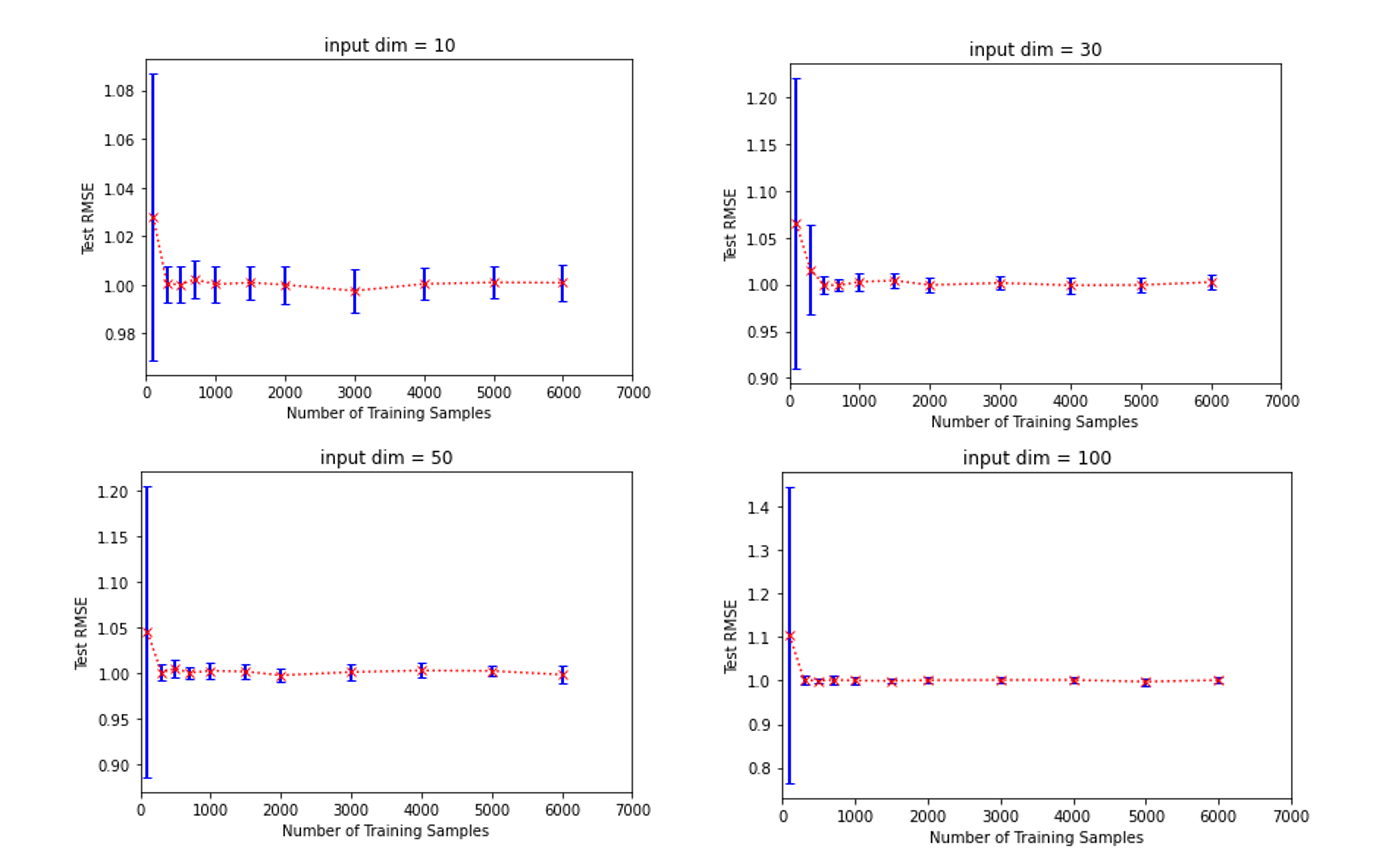}
   \caption{Generalization errors of simulated data sets with $d=10$(top left), $d=30$ (top right), $d=50$ (bottom left), $d=100$ (bottom right) respectively}
    \label{fig:simulation}
\end{figure}

The generalization performance of our trained DCNN is evaluated in terms of the test RMSE (Root Mean Square Error). The experimental results are presented in Figure \ref{fig:simulation}. From Figure \ref{fig:simulation}, we observe that for each $d$, the test RMSE gradually decreases and converges to some constant as the number of training samples $n$ increases. Also, the variance of the test RMSE, as indicated by the blue error bars, reduces significantly as $n$ grows. This verifies our results in Theorem \ref{MainResult2} that $\varepsilon(\pi_M f_{D,J,S}) \rightarrow \varepsilon(f_\rho)$ with high probability.

\section{Related Work, Discussions, and Conclusions}
\label{sec:discussion}
In this section, we discuss prior work in overparameterization in regression and the generalization ability of DCNNs.

As mentioned before, it is frequently observed that overparameterized neural networks generalize well with zero training error for regression loss. Such a phenomenon is known as benign overfitting.
Benign overfitting is characterized in \cite{BartlettBenign} in linear least squares regression with Gaussian data and noise.
Sufficient and necessary conditions are presented for the input covariance matrix's eigenvalue patterns for the minimum-norm interpolator to generalize well.
The methodology there depends heavily on the linearity of the algorithm and the nice properties of the Gaussian random matrices induced by the Gaussian model.
This methodology was extended to a setting of two-layer linear neural networks in \cite{chatterji2022interplay}.
Benign overfitting is also verified for stochastic gradient descent in an overparameterized linear regression setting in \cite{zou2021benign}.
This study was further extended in \cite{cao2022benign} to train a shallow neural network with shared weights by gradient descent with respect to cross-entropy loss. Such a network is equipped with polynomial ReLU activation function $\sigma^q (u) =\max \{0,u\}^q$ with $q>2$:
$$ \frac{1}{m} {\bf 1}_m^T  \left[\sigma^q \left({\bf W}_{+1} y {\bm \mu}\right) + \sigma^q \left({\bf W}_{+1}  {\bm \xi}\right) - \sigma^q \left({\bf W}_{-1} y {\bm \mu}\right) - \sigma^q \left({\bf W}_{-1}  {\bm \xi}\right)\right]. $$
Here, ${\bf W}_{+1}$ and ${\bf W}_{-1}$ are $m\times d$ fully-connected weight matrices shared by a signal input $y {\bm \mu}$, with $y\in \{1, -1\}$ and a fixed vector ${\bm \mu}\in\RR^d$, and a noise vector ${\bm \xi} \in \RR^d$. When ${\bm \xi}$ follows a standard normal distribution $N(0, \sigma_p^2 I_{d-1})$ on the orthogonal complement of ${\bm \mu}$, it was shown in \cite{cao2022benign} that the gradient descent algorithm may achieve an arbitrarily small training error $\epsilon>0$ when the number of iterations is large enough, depending on the accuracy $\epsilon$. 

In a non-linear setting induced by ReLU, benign overfitting is verified for deep fully-connected neural networks in \cite{lin2021generalization}.
This result was achieved by a network deepening scheme where great capacity is provided by the fully-connected structure of the networks to induce interpolations.
In our setting, the convolutional structure of DCNNs gives rigid constraints, making it infeasible to apply the network deepening scheme illustrated in \cite{lin2021generalization}.
We introduce a novel network deepening scheme designated for DCNNs.
We double the width of a given teacher DCNN so that the linear features and the induced hat functions are built for attaining interpolation while the learning ability of the teacher DCNN is preserved.
This is our first novelty. We would like to point out that our results theoretically verify that overparameterized DCNNs can generalize well, but we are not able to characterize benign overfitting of DCNNs. In other words, we are not able to identify a sufficient and necessary condition for DCNNs to achieve benign overfitting. 

Despite the wide applications of DCNNs in modern learning tasks, theoretical understanding of their generalization ability has been challenging.
Promising progress has been made recently.
An estimate of the approximation error of DCNNs is given in \cite{zhou2020universality}.
It shows that the approximation error decreases as the network depth increases.
This result is an important tool we use to prove Theorem \ref{MainResult2}.
In \cite{LinWangWangZhou2022}, it is shown that implementing ERM on DCNNs yields universally consistent estimators without any restrictions on free parameters.
However, an explicit rate of convergence is not provided there.
Moreover, for learning a radial regression function by a DCNN followed by one fully-connected layer, a learning rate of order $\mathcal{O}(n^{-1/2})$ is obtained in \cite{mao2021theory}.
This learning rate is derived using a covering number estimate, which requires the filters and biases to be bounded, with bounds depending on the layers where the parameters lie. 
More recently, for learning an additive ridge function by a DCNN followed by one fully-connected layer, a learning rate of order $\mathcal{O}\left(n^{\frac{-2\alpha}{1+\alpha}}\log n\right)$ is obtained in \cite{fang2022optimal}, where $0 < \alpha \leq 1$. This learning rate is considered to be optimal up to a log factor. This result also requires the filters and bias to be bounded.

Some neural networks with similar structures are studied in the literature.
For example, periodized CNNs are studied in \cite{Petersen2020} where convolutions on sequences on the group $\ZZ_d$ are used to induce circular weight matrices instead of Toeplitz-type ones induced by regular convolutions on sequences on $\ZZ$.
Properties of approximating  translation
equivariant functions by such neural networks are discussed. 
Other work studying CNN-variant neural networks include ResNet-type CNNs in \cite{Oono2019}, and fully-connected networks inducing sparsity in \cite{Grohs}.

In this paper, we derived, in our belief, the first learning rate of DCNNs that has been presented so far without any restrictions on free parameters or additional fully-connected layers.
Such a breakthrough can be attributed to a novel sample error estimate, which relies on a tight pseudo-dimension estimate of DCNNs with piecewise linear activation functions inspired by that in \cite{Bartlett2019}.
Our second novelty is making use of this upper bound of pseudo-dimension, which no longer requires restrictions on filters and biases and can be used to bound the empirical covering number in turn.

\section{Acknowledgements}
The authors would like to thank the referees for their detailed comments and constructive suggestions that led to
an improved presentation of this paper. The authors are partially sponsored by NSF grants CCF-1740776, DMS 2015363, and IIS-2229876. They are also partially supported by the A. Russell Chandler III Professorship at Georgia Tech.

\bibliographystyle{plain}
\bibliography{cnn}

\section*{Appendix A }
\section*{A.1\quad Proof of Lemma \ref{matrixproducts}} \label{sec:A1}
\begin{proof}
The case $j=1$ is trivial. 
Suppose that (\ref{filterI}) holds for $j=m$ with $m\in\{1, \ldots, J^{*}-1\}$. That is, the $(D +ms) \times D$ matrix $\prod_{i=1}^m T^{(i)} = T^{{\bf w}, m}$ satisfies
$$ \left(T^{(m)} \ldots T^{(1)}\right)_{\ell, k} = \left(T^{{\bf w}, m}\right)_{\ell, k} = W^{(m)}_{\ell -k}, \qquad 1 \leq \ell\leq D +ms, \ 1 \leq k \leq D. $$
Now we consider the product $T^{(m+1)} \left(\prod_{i=1}^m T^{(i)}\right)$, where $T^{(m+1)}$ is a $(D + (m+1)s) \times (D +ms)$ matrix.  The $(i, k)$-entry of this matrix product (with $1 \leq i \leq D +(m+1)s$, $1\leq k \leq D$) equals
\begin{align*}
\left(T^{(m+1)} \left(T^{(m)} \ldots T^{(1)}\right)\right)_{i, k} &= \sum_{\ell =1}^{D +ms} \left(T^{(m+1)}\right)_{i, \ell} \left(T^{(m)} \ldots T^{(1)}\right)_{\ell, k} \\
&= \sum_{\ell =1}^{D +ms} \left(w^{(m+1)}\right)_{i -\ell} W^{(m)}_{\ell -k}.
\end{align*}
Note that the sequence $W^{(m)}$ is  supported in $\{0, \ldots, m s\}$. For $\ell \leq 0$, we have $\ell - k <0$ and thereby $W^{(m)}_{\ell -k}=0$. 
For $\ell \geq d_{m} +1$, since $k \leq d$, we have $\ell - k \geq  ms +1$ implying $W^{(m)}_{\ell -k}=0$. So there holds 
$$\left(T^{(m+1)} \left(T^{(m)} \ldots T^{(1)}\right)\right)_{i, k}
= \sum_{\ell =-\infty}^{\infty} \left(w^{(m+1)}\right)_{i -\ell} W^{(m)}_{\ell -k}. $$
This is exactly the $(i-k)$-th entry of the convoluted sequence $w^{(m+1)}{*}W^{(m)}$. This is also the $(i, k)$-entry of the $(D +(m+1)s) \times D$ matrix $T^{{\bf w}, m+1}$. This proves (\ref{filterI}) for $j=m+1$, which completes the induction procedure and the proof of the lemma. 
\end{proof}

\section*{A.2 \quad Proof of Theorem \ref{generalpseudodim}: Pseudo-dimension of feed-forward neural networks} \label{sec:A.2}

To prove Theorem \ref{generalpseudodim}, we need the following technical tool on the number of possible sign vectors attained by polynomial vectors. It is introduced in \cite[Lemma 2.1]{bartlett1998almost} (the proof can be found in \cite[Theorem 8.3]{anthony1999neural}).

\begin{lemma}\label{polysignlemma}
Let $f_1, \ldots, f_\ell$ be polynomials of degree at most $L$ in $\kappa \leq \ell$ variable. Then the number of distinct sign vectors $\{\hbox{sgn} (f_1 (a)), \ldots, \hbox{sgn}(f_\ell (a))\} \in \{1, 0\}^\ell$ that can be generated by varying $a\in\RR^\kappa$ is at most $2(2 e \ell L/\kappa)^\kappa$. Here $\hbox{sgn} (u) =1$ if $u\geq 0$ and $0$ otherwise. 
\end{lemma}

\begin{proof}[Proof of Theorem \ref{generalpseudodim}] 
For an arbitrary choice of $\ell$ points $x_1,\ldots,x_\ell \in \RR^d$, we wish to bound 
\begin{equation}\label{signnumberdef}
\mathcal{K}:=|\left\{(\text{sgn}(f(x_1,a)), \ldots \text{sgn}(f(x_\ell,a))): a\in \RR^{K}\right\}|.
\end{equation}
In other words, $\mathcal{K}$ is the number of sign patterns that the neural network can output for
the sequence of inputs $(x_1, \ldots, x_\ell)$. We will derive upper bounds for $\mathcal{K}$, leading us to the pseudo-dimension estimate for $\mathcal{H}$. To do so, we partition the free parameter domain $\RR^K$ into some non-overlapping subsets in such a way that within each subset, the functions $f(x_1, \cdot), \ldots, f(x_\ell, \cdot)$ are fixed polynomials of degree at most $1+\theta+\cdots+\theta^{J}$. Fix $\{x_1,\ldots,x_\ell\} \subset \RR^d$. Let $K \leq \ell$. 

We will construct the desired partition by defining a sequence of partitions $\{{\mathcal S}^{(i)}\}_{i=0}^J$ of $\RR^K$ iteratively layer by layer with ${\mathcal S}^{(0)} =\{\RR^K\}$ such that:
\begin{enumerate}
    \item For each $i=1, \ldots, J$,
    \begin{equation}\label{ratioSi}
\frac{\left|{\mathcal S}^{(i)}\right|}{\left|{\mathcal S}^{(i-1)}\right|} \leq N_i :=2\left(\frac{2 e d_i \ell p(1 + \theta+\theta^2 + \cdots+\theta^{i-1})}{K_1 + \cdots +K_i}\right)^{K_1 + \cdots +K_i}.
\end{equation}
\item For each $i=1, \ldots, J+1$, on each element $S$ of ${\mathcal S}^{(i-1)}$, for each $j\in\{1, \ldots, \ell\}$ and $k\in \{1, \ldots, d_i\}$, the net input component $$\left(T^{(i)} h^{(i-1)}(x_j, a)  - b^{(i)}\right)_k$$ is a polynomial of degree at most $1 + \theta+\theta^2 + \cdots+\theta^{i-1}$ of the variables $A^{(1)}, \ldots, A^{(i)}$. Here for $i=J+1$ with $d_{J+1}=1, T^{(J+1)} = c^T, A^{(J+1)} =c$, the only input component is the output function value $f(x_j, a) =c\cdot  h^{(J)} (x_j, a)$ at $x_j$ for each $j\in\{1, \ldots, \ell\}$.
\end{enumerate}

Note that $|{\mathcal S}^{(0)}|=1$. For $i=1$, the above properties for ${\mathcal S}^{(1)}$ holds true because each input component 
$$\left(T^{(1)} h^{(0)}(x_j, a)  - b^{(1)}\right)_k = \left(T^{(1)} x_j  - b^{(1)}\right)_k$$ is an affine function of the variables $A^{(1)}$. 

Suppose that the partitions $\{{\mathcal S}^{(i)}\}_{i=0}^{q-1}$ have been found. Let us construct ${\mathcal S}^{(q)}$. By our induction hypothesis, for each $j\in\{1, \ldots, \ell\}$ and $k\in \{1, \ldots, d_{q}\}$, the input component $\left(T^{(q)} h^{(q-1)}(x_j, a)  - b^{(q-1)}\right)_k$ restricted onto each element $S$ of ${\mathcal S}^{(q-1)}$ equals to a polynomial, denoted as $P_{k, j, S}(a)$, of degree at most $1 + \theta+\theta^2 + \cdots+\theta^{q-1}$ of the variables $A^{(1)}, \ldots, A^{(q)}$. 
For $S \in {\mathcal S}^{(q-1)}$, we consider a collection of polynomials 
$$\left\{P_{k, j, S}(a)-t_s: k\in \{1, \ldots, d_{q}\},  j\in\{1, \ldots, \ell\}, s\in\{1, \ldots, p\}\right\}. $$
Since the number of variables $K_1 + \ldots + K_q$ in $A^{(1)}, \ldots, A^{(q)}$ is bounded by the number of functions $d_{q}\ell p$ in the collection, applying Lemma \ref{polysignlemma}, we know that the number of distinct sign vectors achieved 
by this collection is at most $N_q$. 
Hence we can partition $\RR^K$ into at most $N_q$ non-overlapping subsets such that all these polynomials keep the signs unchanged on each subset. Intersecting these subsets with $S$ gives a partition of $S$ into at most $N_q$ non-overlapping subsets. These partitions with $S$ running over ${\mathcal S}^{(q-1)}$ form a partition of $\RR^K$ which is a refinement of  ${\mathcal S}^{(q-1)}$ and is the desired partition ${\mathcal S}^{(q)}$. Obviously, (\ref{ratioSi}) is satisfied for $i=q$. Moreover, on each $S' \in {\mathcal S}^{(q)}$ in the partition of $S \in {\mathcal S}^{(q-1)}$, for each $j\in\{1, \ldots, \ell\}$ and $k\in \{1, \ldots, d_{q}\}$, the polynomials $P_{k, j, S}(a)-t_s$ keep the signs unchanged and thereby for all $a\in S'$, the value $P_{k, j, S}(a)$ lies on the same interval $[t_{\alpha}, t_{\alpha+1})$ for some $\alpha\in \{0, \ldots, p\}$.  In fact, $\alpha$ is either the maximum $s\in \{1, \ldots, p\}$ such that $\hbox{sgn} (P_{k, j, S}(a)-t_s) =1$ or $0$ if $\hbox{sgn} (P_{k, j, S}(a)-t_s) =0$ for all $s\in \{1, \ldots, p+1\}$. 

As $\sigma|_{[t_{\alpha }, t_{\alpha +1})}$ is a polynomial of degree at most $\theta$, we know that on $S' \subseteq S$, each component of $h^{(q)}(x_j, a)$, with $k\in\{1, \ldots, d_q\}$, 
$$  \left(h^{(q)}(x_j, a)\right)_k = \sigma\left(\left(T^{(q)} h^{(q-1)}(x_j, a)  - b^{(q-1)}\right)_k\right) = \sigma\left(P_{k, j, S}(a)\right) $$ 
is a polynomial of degree at most $\theta (1 + \theta+\theta^2 + \cdots+\theta^{q-1})$. This implies that each input component $\left(T^{(q+1)} h^{(q)}(x_j, a)  - b^{(q+1)}\right)_k$ is a polynomial of degree at most $1 + \theta+\theta^2 + \cdots+\theta^{q}$ of the variables $A^{(1)}, \ldots, A^{(q+1)}$ on $S' \in {\mathcal S}^{(q)}$. This verifies the desired property for ${\mathcal S}^{(q)}$ and completes the induction procedure. 

Thus, we have constructed a partition ${\mathcal S}^{(J)}$ of $\RR^K$ such that on each element $S$ of ${\mathcal S}^{(J)}$, for each $j\in\{1, \ldots, \ell\}$, the function $f(x_j, a) =c\cdot  h^{(J)} (x_j, a)$ is a polynomial of degree at most $1 + \theta+\theta^2 + \cdots+\theta^{J}$ of the variables $a=[A^{(1)}, \ldots, A^{(J)}, c]$. Since $\ell \geq K$, applying Lemma \ref{polysignlemma}, we know that 
$$ |\left\{(\text{sgn}(f(x_1,a)), \ldots \text{sgn}(f(x_\ell,a))): a\in S\right\}| \leq 2 \left(\frac{2 e \ell (1 + \theta+\theta^2 + \cdots+\theta^{J})}{K}\right)^K. $$
Since ${\mathcal S}^{(J)}$ is a partition of $\RR^K$, we see that the number of sign patterns $\mathcal{K}$ defined by (\ref{signnumberdef}) can be bounded as 
\begin{eqnarray*}
 \mathcal{K} &\leq& \left|{\mathcal S}^{(J)}\right| 
2 \left(2 e \ell (1 + \theta+\theta^2 + \cdots+\theta^{J})/K\right)^K \\
&\leq& \left(\Pi_{i=1}^J N_i\right)
2 \left(2 e \ell (1 + \theta+\theta^2 + \cdots+\theta^{J})/K\right)^K. 
\end{eqnarray*}
Bounding the geometric mean by the arithmic one,  with $W_i :=K_1 + \cdots +K_i$, we find 
$$ \mathcal{K} \leq 2^{J+1} \left(\frac{2 e \ell \left(1 + \theta+\theta^2 + \cdots+\theta^{J} 
+ \sum_{i=1}^J d_i p (1 + \theta+\theta^2 + \cdots+\theta^{i-1})\right)}{K +\sum_{i=1}^J W_i}\right)^{K +\sum_{i=1}^J W_i}.  
$$
Note that the points $\{x_1, \ldots, x_\ell\}$ are arbitrarily chosen. An upper bound for the VC-dimension is then obtained by computing the largest value of $\ell$ for which  
the above number is at least $2^\ell$, yielding
$$
2^\ell \leq 2^{J+1} \left(\frac{2 e \ell \left(1 + \beta+\beta^2 + \cdots+\beta^{J} 
+ \sum_{i=1}^J d_i p (1 + \beta+\beta^2 + \cdots+\beta^{i-1})\right)}{K +\sum_{i=1}^J W_i}\right)^{K +\sum_{i=1}^J W_i}.  
$$
Then the desired bound follows by applying Lemma \ref{lemma18} below and noticing that 
$$K +\sum_{i=1}^J W_i 
=d_J + \sum_{j=1}^J K_j +\sum_{i=1}^J \sum_{j=1}^i K_j  =d_J + \sum_{j=1}^J (J- j +2) K_j \geq K. $$
This completes the proof of Theorem \ref{generalpseudodim}. 
\end{proof}

\begin{lemma}[Lemma 18 in \cite{Bartlett2019}] \label{lemma18}
Suppose that $2^m \leq 2^t(mr/w)^w$ for some $r\geq 16$ and $m\geq w \geq t \geq 0$. Then, $m\leq t+w \log_2(2r\log_2 r)$.
\end{lemma}

\section*{A.3 \quad Proof of Lemma \ref{Lemma:Pseudo-for-DCNN} }\label{sec:A.3}

\begin{proof}
For neural network consider in Lemma \ref{Lemma:Pseudo-for-DCNN}, we have $S+1$ free parameters in each $w^{(j)}$, and $2S-1$  free parameters from $b^{(j)}$ in the $j$-th layers for $j=1,\dots,J-1$ and $S+1+d+JS$ free parameters in the $J$-th layer. In other words, we have 
$$ K_j = 3 S, \qquad \text{for }j=1, \ldots, J-1$$
and
$$ K_J= S+1+d+JS. $$
Then, we have 
\begin{align*}
d_J + \sum_{j=1}^J (J- j +2) K_j &= d+JS +\sum_{j=1}^{J-1}(J- j +2) K_j + 2K_J\\
&=2S+2 + 3(d+JS) + \frac{3}{2}S(J+4)(J-1)\\
&= 2 + 3d -4S +\frac{15}{2}JS  + \frac{3}{2}J^2S.
\end{align*}
For ReLU, we have $\theta = 1 $ and $p=1$. Then, \begin{align*}
R&=1 + J
+ \sum_{i=1}^J (d + iS) i  \\
&= 1 + J
+ \frac{J(J+1)}{2} d + \frac{J(J+1)(2J+1)}{6} S\\ &=\frac{1}{3} J^3 S + \frac{1}{2} J^2 (d+S) + J \left(\frac{d}{2} + \frac{S}{6} +1\right) +1 . 
\end{align*}
Observe that $$ 2 + 3d -4S + \frac{15}{2}JS  + \frac{3}{2}J^2S \leq 3d  + 9 J^2S \qquad (\because S \geq 2, J^2 \geq J)$$and 
$$ R \leq \frac{1}{3} J^3 S + J^2 (d+S+1) +1 \leq 3 J^3S+ J^2d \leq 3(J^2 S + Jd)^2. $$
Hence
\begin{align*}
\text{Pdim}(\mathcal{H}_{J, S})
&\leq J+1 + (3d  + 9 J^2S) \left(\log_2 (12 e (J^2 S + Jd)^2) +  \log_2 \log_2 (6 e (J^2 S + Jd)^2)\right)\\
&\leq J+1 + (3d  + 9 J^2S) 2\left(\log_2 (12 e (J^2 S + Jd)^2)\right) \\
&\leq c_0(J^2S+d)\log(Jd+J^2S),
\end{align*}
where $c_0$ is an absolute constant. The proof is complete. 
\end{proof}

\end{document}